\documentclass[10pt,twocolumn,letterpaper]{article}

\usepackage{titling}
\usepackage[pagenumbers]{cvpr} 
\usepackage[toc,page,header]{appendix}
\usepackage{minitoc}

\usepackage{graphicx}
\graphicspath{{./figures/}}
\usepackage{comment}
\usepackage{amsmath,amssymb} 
\usepackage{color}
\usepackage{booktabs}
\usepackage[normalem]{ulem}
\useunder{\uline}{\ul}{}
\usepackage{multirow}
\usepackage{xspace}
\usepackage{xcolor}
\usepackage{cite}
\usepackage{adjustbox}
\usepackage{wrapfig}
\usepackage{placeins}
\usepackage[htt]{hyphenat}
\usepackage{enumitem}
\usepackage{pifont}
\newcommand{\cmark}{\ding{51}}%
\newcommand{\xmark}{\ding{55}}%
\usepackage[symbol]{footmisc}

\usepackage[pagebackref,breaklinks,colorlinks]{hyperref}
\hypersetup{ colorlinks=true, linkcolor=blue, filecolor=magenta, urlcolor=cyan, }

\usepackage[accsupp]{axessibility}  

\usepackage[capitalize]{cleveref}
\crefname{section}{Sec.}{Secs.}
\Crefname{section}{Section}{Sections}
\Crefname{table}{Table}{Tables}
\crefname{table}{Tab.}{Tabs.}

\definecolor{rgb_red}{HTML}{BB3E03}
\definecolor{depth_yellow}{HTML}{EE9B00}
\definecolor{semantic_teal}{HTML}{0A9396}
\definecolor{deemph}{gray}{0.6}
\newcommand{\gc}[1]{\textcolor{deemph}{#1}}

\newcommand\crgb[1]{\textcolor{rgb_red}{#1}}
\newcommand\cdepth[1]{\textcolor{depth_yellow}{#1}}
\newcommand\csemseg[1]{\textcolor{semantic_teal}{#1}}

\newcommand\rgb{\crgb{RGB}\xspace}
\newcommand\depth{\cdepth{D}\xspace}
\newcommand\semseg{\csemseg{S}\xspace}
\newcommand\pdepth{\cdepth{pD}\xspace}

\newcommand\rgbd{\crgb{RGB}-\cdepth{D}\xspace}
\newcommand\rgbs{\crgb{RGB}-\csemseg{S}\xspace}
\newcommand\depthsemseg{\cdepth{D}-\csemseg{S}\xspace}
\newcommand\rgbpd{\crgb{RGB}-\cdepth{pD}\xspace}
\newcommand\rgbps{\crgb{RGB}-\csemseg{pS}\xspace}
\newcommand\rgbds{\crgb{RGB}-\cdepth{D}-\csemseg{S}\xspace}
\newcommand\rgbpdps{\crgb{RGB}-\cdepth{pD}-\csemseg{pS}\xspace}

\newcommand\mmae{\mbox{MultiMAE}\xspace}

\def\cvprPaperID{*****} 
\def\confName{CVPR}
\def\confYear{2022}

\begin{document}
\doparttoc 
\faketableofcontents 

\title{\mmae: Multi-modal Multi-task Masked Autoencoders \vspace{-4mm}}

\author{Roman Bachmann\footnotemark[1] \quad\quad David Mizrahi\footnotemark[1] \quad\quad Andrei Atanov \quad\quad Amir Zamir \\ 
Swiss Federal Institute of Technology Lausanne (EPFL) \vspace{3mm}\\
\url{https://multimae.epfl.ch}
}

\twocolumn [{%
    \renewcommand\twocolumn[1][]{#1}%
    \vspace{-10mm}
    \maketitle
    \centering
    \includegraphics[width=1.0\textwidth]{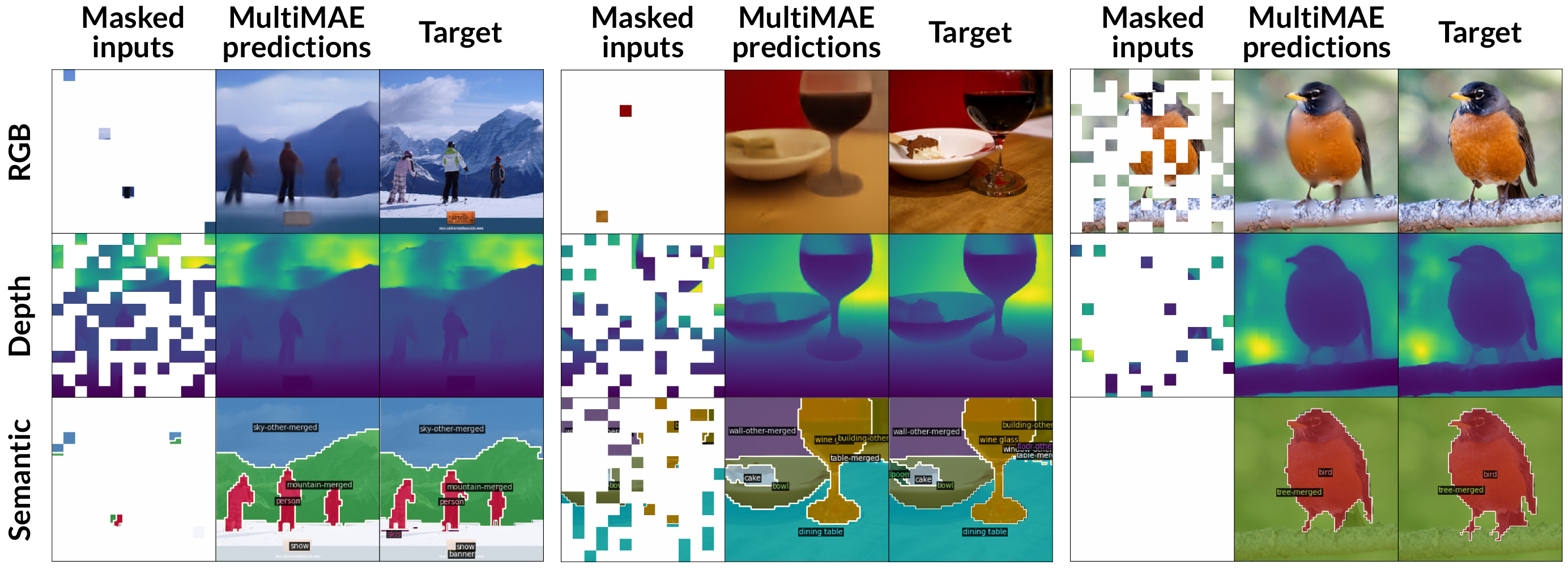}
    \captionof{figure}{
        \textbf{\mmae pre-training objective.}
        We randomly select 1/6 of all 16$\times$16 image patches from multiple modalities and learn to reconstruct the remaining 5/6 masked patches from them. The figure shows validation examples from ImageNet, where masked inputs (left), predictions (middle), and non-masked images (right) for \crgb{RGB} (top), \cdepth{depth} (middle), and \csemseg{semantic segmentation} (bottom) are provided.
        Since we do not compute a loss on non-masked patches, we overlay the input patches on the predictions.
    }
    \label{fig:random_samples}
    \vspace{5mm}
}]

\footnotetext[1]{Equal contribution.}


\begin{abstract}
    \vspace{-2.0mm}
We propose a pre-training strategy called Multi-modal Multi-task Masked Autoencoders (\mmae). It differs from standard Masked Autoencoding in two key aspects: \textbf{I)} it can \textbf{optionally} accept additional modalities of information in the input besides the RGB image (hence ``multi-modal"), and \textbf{II)} its training objective accordingly includes predicting multiple outputs besides the RGB image (hence ``multi-task"). 

We make use of masking (across image patches and input modalities) to make training \mmae \textbf{tractable} as well as to ensure \textbf{cross-modality predictive coding} is indeed learned by the network.
We show this pre-training strategy leads to a flexible, simple, and efficient framework with improved transfer results to downstream tasks. In particular, the same exact pre-trained network can be flexibly used when additional information besides RGB images is available or when no information other than RGB is available – in all configurations yielding competitive to or significantly better results than the baselines. 
To avoid needing training datasets with multiple modalities and tasks, we train \mmae \textbf{entirely using pseudo labeling}, which makes the framework widely applicable to any RGB dataset. 

The experiments are performed on multiple transfer tasks (image classification, semantic segmentation, depth estimation) and datasets (ImageNet, ADE20K, Taskonomy, Hypersim, NYUv2). The results show an intriguingly impressive capability by the model in cross-modal/task predictive coding and transfer. 
Code, pre-trained models, and interactive visualizations are available at \url{https://multimae.epfl.ch}.
    \vspace{-3.0mm}
\end{abstract}



\section{Introduction}

Masked Autoencoders (MAEs)~\cite{He2021MAE} have recently been demonstrated to be a powerful, yet conceptually simple and efficient, self-supervised pre-training strategy for Vision Transformers~\cite{Dosovitskiy2021ViT} (ViTs). 
Their training objective is to mask-out a high number of patches in an input image and to predict the missing regions. 
To that end, only the small number of non-masked patches are first processed using a Transformer encoder~\cite{Vaswani2017Attention}, and then decoded with a light-weight Transformer that reconstructs the original image.
To solve this task sufficiently well, it is assumed~\cite{He2021MAE} that the network needs to learn representations that capture more than just low-level image statistics. 

So far, however, the MAE pre-training objective has been limited to a single modality, namely RGB images, and does not make use of any other modalities that are optionally present. In practice, often more than only a single modality of information is available, either through sensing (e.g., a depth sensor) or pseudo labeling (e.g., a powerful pre-trained depth estimation network). Multi-modality is also argued to be employed by biological organisms to develop resilience and better representations ~\cite{smith2005development,de2003sensory,de1998category}.
As we demonstrate in our experiments, making use of such optionally present modalities has the potential to greatly improve the performance of downstream tasks, compared to using only RGB images.

Besides multi-modality (i.e., different inputs), multi-taskness (i.e., different outputs) is an important aspect, as it has been shown that there is usually no single pre-training objective that transfers best to all possible downstream tasks~\cite{Zamir2018Taskonomy, Sax2018MidLevel, Mensink2021Factors}.
Instead, pre-training with a diverse set of tasks~\cite{Baxter2000InductiveBias, Tripuraneni2020TaskDiversity} has been observed to improve the performance on downstream tasks~\cite{Tian2020RethinkingFewShot, Ghiasi2021MuST} and potentially learn a better representation. In general, modifying the training objectives is a powerful way to steer what representation the model will learn.

\begin{figure*}[t]
\includegraphics[width=1.0\textwidth]{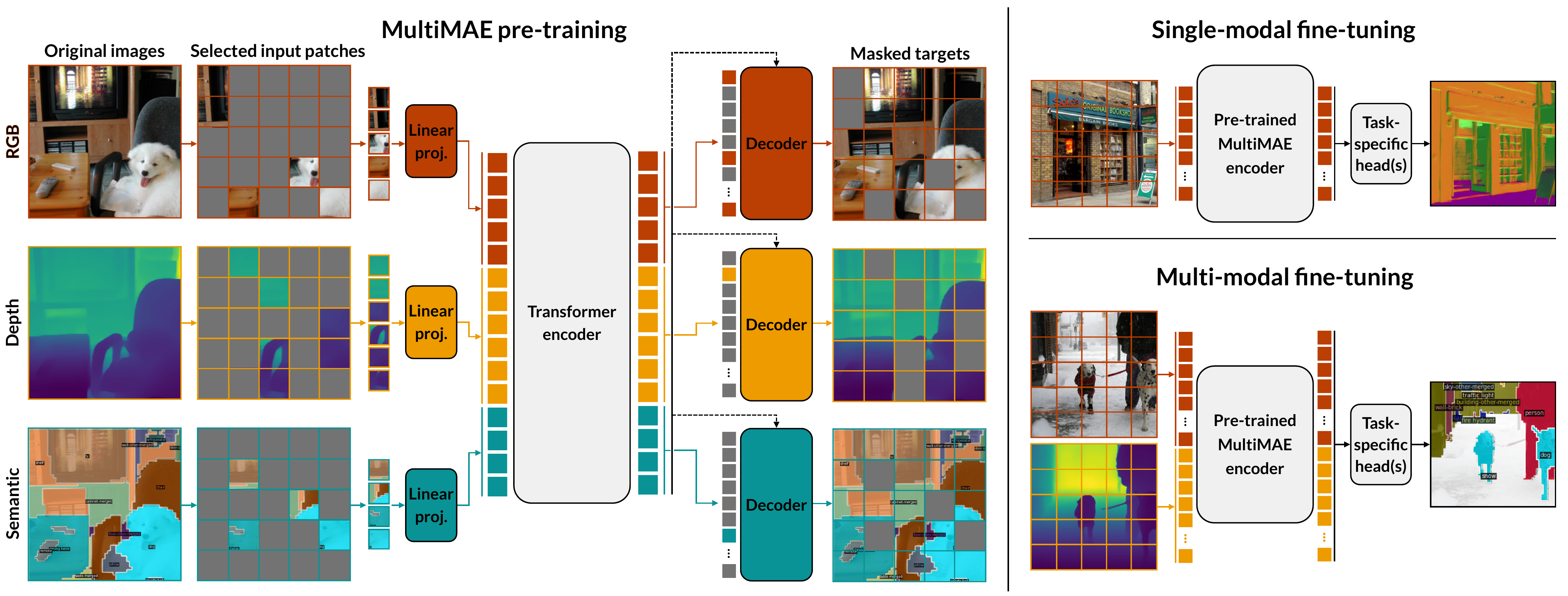}
\caption{
    \textbf{(Left) \mmae pre-training}: 
    A small subset of randomly sampled patches from multiple modalities (e.g., \rgb, \cdepth{depth}, and \csemseg{semantic segmentation}) is linearly projected to tokens with a fixed dimension and encoded using a Transformer. 
    Task-specific decoders reconstruct the masked-out patches by first performing a cross-attention step from queries to the encoded tokens, followed by a shallow Transformer.
    The queries consist of mask tokens (in gray), with the task-specific encoded tokens added at their respective positions.
    \textbf{(Right) Fine-tuning}: 
    By pre-training on multiple modalities, \mmae lends itself to fine-tuning on single-modal and multi-modal downstream tasks.
    No masking is performed at transfer time.
}
\label{fig:method_figure}
\centering
\vspace{-1em}
\end{figure*}

In this paper, we present Multi-modal Multi-task Masked Autoencoders (\mmae), a simple and effective method to make masked autoencoding include multiple modalities and tasks (see Fig.~\ref{fig:method_figure}). 
In particular, in our current instantiation of this general method, we study adding dense scene depth to capture geometric information, as well as segmentation maps to include information about the semantic content of the scene. We created a multi-task dataset by pseudo labeling these tasks on ImageNet-1K~\cite{Deng2009ImageNet,Ghiasi2021MuST}.
This has the advantage that in order to train a \mmae, one only requires a large unstructured RGB dataset without annotations and only off-the-shelf neural networks to perform the pseudo labeling.

To train \mmae, we randomly sample a small set of patches from different input modalities, and encode them using a Transformer encoder.
\mmae's objective is then to reconstruct the masked-out patches of all tasks using task-specific decoders.
Figure~\ref{fig:random_samples} shows example predictions for the multi-task masked reconstruction that \mmae performs.
\mmae has to learn not only the original MAE objective (within-RGB in-painting), but also to reconstruct any task from any input modality (cross-modal prediction) all from a very sparse set of input patches.
The first objective leads to learning \emph{spatial predictive coding} while the second one leads to \emph{cross-modal predictive coding}. 

\section{Related Work}

\noindent\textbf{Masked image prediction} consists of learning useful representations by learning to reconstruct images corrupted by masking. This approach was pioneered with denoising autoencoders~\cite{vincent_stacked_2010} and context encoders~\cite{pathak2016context}. 
With the introduction of Vision Transformers (ViT)~\cite{Dosovitskiy2021ViT} and motivated by the success of BERT~\cite{devlin2018bert} in NLP, many recent works propose a variety of masked image prediction methods for pre-training vision models in a self-supervised way, using reconstruction targets such as pixels \cite{chen_generative_2020, Dosovitskiy2021ViT, atito2021sit, He2021MAE, Xie2021SimMIM, el2021large}, discrete tokens \cite{Bao2021BEiT, zhou2021ibot}, and (deep) features ~\cite{wei2021masked, baevski2022data2vec}. These methods scale very well and achieve strong results on various downstream tasks including motor control~\cite{Xiao2022}. In particular, the masked autoencoder (MAE)~\cite{He2021MAE} approach accelerates pre-training by using an asymmetric architecture consisting of a large encoder that operates \textit{only} on unmasked patches followed by a lightweight decoder that reconstructs the masked patches from the latent representation and mask tokens. Our approach leverages the efficiency of the MAE approach and extends it to multi-modal and multi-task settings.

\smallskip
\noindent\textbf{Multi-modal learning} involves building models capable of relating information from multiple sources. 
It can either involve training separate encoders or one unified architecture (e.g., a Transformer \cite{Vaswani2017Attention}) to operate on modalities such as images and text \cite{castrejon2016learning, kaiser2017one, karpathy2015deep, lu202012, alayrac2020self, chen2020uniter, tan2019lxmert,  hu2021unit, kamath2021mdetr, lu2019vilbert, su2019vl, kim2021vilt, xu2021e2e}, video and audio \cite{arandjelovic2017look, owens2018audio, nagrani2021attention, jaegle2021perceiver}, video, text and audio \cite{akbari2021vatt}, and depth, images and video \cite{girdhar2022omnivore}.
Our work proposes a simple approach to pre-train Transformers on multiple dense visual modalities and produce strong cross-modal interaction. Unlike most prior work which assumes that all modalities are available during inference, our approach is designed to perform well on any subset of the pre-training modalities.

Related to \mmae are several works that perform multi-modal autoencoding~\cite{Ngiam2011MultimodalDL, Wu2018MultimodalGM, Shi2019VariationalMA, Sutter2019MultimodalGL, Sutter2021GenELBO}.
Our approach differs from them in that we use a more flexible architecture and perform masked autoencoding to learn cross-modal predictive coding among optional inputs (as demonstrated in Fig.~\ref{fig:random_samples}).

\smallskip
\noindent\textbf{Multi-task learning} consists of training models to predict multiple output domains from a single input~\cite{caruana_multitask_1997, eigen2015predicting, kokkinos2017ubernet}. In computer vision, the input is usually an RGB image. A common approach for multi-task learning is to use a single encoder to learn a shared representation followed by multiple task-specific decoders~\cite{vandenhende2021multi, Ghiasi2021MuST}. These methods differ from our approach as we use multiple tasks in both the input and the output along with masking.

In addition, many works study the importance of task diversity to improve transfer performance~\cite{Zamir2018Taskonomy, Sax2018MidLevel, Ghiasi2021MuST, Mensink2021Factors, tripuraneni2020theory}. These works argue that learning from one task alone is insufficient and that a set of tasks can more effectively cover the many possible downstream tasks in vision. Our pre-training method operates on multiple tasks to learn more general representations capable of covering multiple downstream tasks.

\smallskip
\noindent\textbf{Self-training} is a technique to incorporate unlabeled data into a supervised learning setting~\cite{yarowsky1995unsupervised, scudder1965probability,rosenberg2005semi, lee2013pseudo}. It is one of the earliest approaches to semi-supervised learning.
Self-training methods use a supervised model to generate pseudo labels on unlabeled data and then train a student model on the pseudo labeled data. These approaches have been applied to a variety of vision tasks such as image classification~\cite{yalniz2019billion, xie2020self, pham2021meta}, object detection~\cite{zoph2020rethinking}, and segmentation~\cite{chen2020naive, zoph2020rethinking}. Most recently, multi-task self-training (MuST)~\cite{Ghiasi2021MuST} uses specialized teachers to create a multi-task pseudo labeled dataset and then trains a multi-task student model on this dataset to learn general feature representations. Our method also relies on pseudo labeling to produce a large-scale multi-task dataset. However, unlike prior work, pseudo labels are not only used as output targets but also as \emph{masked} input modalities. 

\section{Method Description}
In this Section, we describe the Multi-modal Multi-task Masked Autoencoder (\mmae) architecture (illustrated in Fig.~\ref{fig:method_figure}), as well as the pre-training strategy in more detail.
We first give an architectural overview of both the multi-modal encoder (Sec.~\ref{sec:multimodal_encoder}) and multi-task decoders (Sec.~\ref{sec:decoders}).
We then describe our multi-modal token sampling strategy (Sec.~\ref{sec:multimodal_masking}) and introduce the pseudo labeled tasks we use for pre-training (Sec.~\ref{sec:pseudo_labeling}).
Finally, we display the most important pre-training details (Sec.~\ref{sec:pretraining_details}).

\subsection{Multi-modal encoder} \label{sec:multimodal_encoder}
Our multi-modal Transformer encoder is a ViT~\cite{Dosovitskiy2021ViT}, but with patch projection layers for each additional input modality.
Specifically, 16$\times$16 patches of each modality are projected to tokens with the correct Transformer dimension using a different linear projection for each modality.
Projected patches are concatenated into a sequence of tokens and given as input to the same Transformer encoder. 
We also add an additional \emph{global} token with a learned embedding, similar to the class-token used in ViT.
Due to the architectural similarities to ViT, \mmae pre-trained weights can directly be used in a standard single-modal ViT by loading only the desired input projection and ignoring the others.

\smallskip
\noindent\textbf{Positional, modality and class embeddings.} 
Since all our modalities have a 2D structure, we add 2D sine-cosine positional embeddings~\cite{Chen2021MoCoV3, He2021MAE} after the linear projection.
We do not explicitly add any modality-specific embeddings, since the bias term in each linear projection can act as such.
In order to perform the semantic segmentation patch projection, we first replace each class index with learned 64-dimensional class embeddings.

\smallskip
\noindent\textbf{Low computational complexity.} 
Just as in the RGB-only MAE~\cite{He2021MAE}, we only pass the small randomly sampled subset of all tokens to the Transformer encoder as part of the masked autoencoding objective.
This is in contrast to the masked autoencoding approaches of SiT~\cite{Ahmed2021SiT}, BeiT~\cite{Bao2021BEiT} and SimMIM~\cite{Xie2021SimMIM}, that encode both the masked and visible tokens.
Due to the quadratic complexity of standard self-attention as a function of the number of tokens, encoding only the random subset of visible tokens becomes increasingly important as the number of input modalities grows.
Indeed, the speedup and reduction in memory are significant and crucial in enabling \mmae's multi-modal pre-training with three dense input modalities. 
A comparison of the pre-training time with and without masked tokens is given in the Appendix (Sec. \ref{sec:runtime_supp}).

\begin{figure*}[t]
\centering
\includegraphics[width=1.0\textwidth]{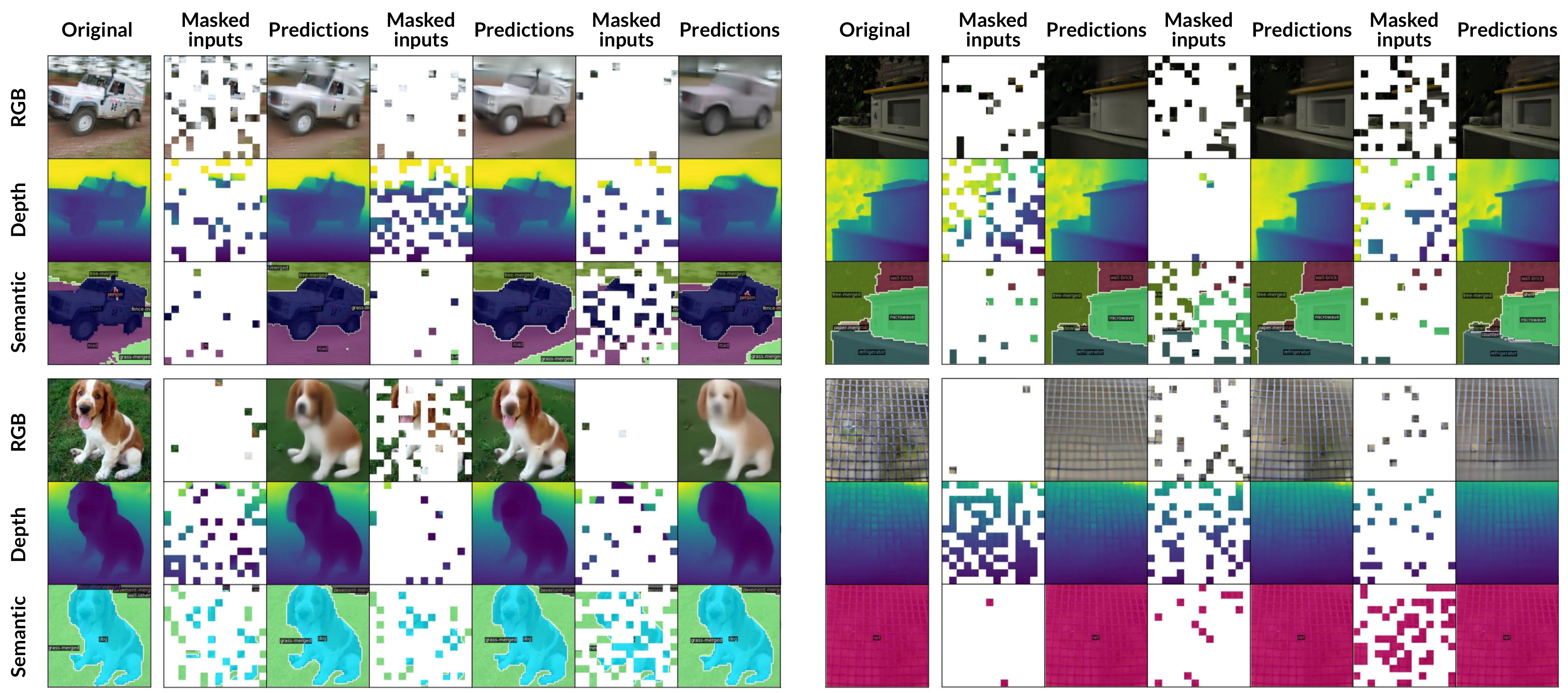}
\caption{
    \textbf{\mmae predictions for several randomly sampled masks.} For each ImageNet validation image, we randomly sample three masks using Dirichlet concentration parameter $\alpha=1$.  Only 1/6 of total patches are left unmasked. Even when very few tokens from one modality are visible, the resulting predictions are relatively stable and plausible due to cross-modal interaction. More examples are shown in the Appendix and on \href{https://multimae.epfl.ch/\#random-masks}{our website}.
    }
\label{fig:dirichlet_samples}
\centering
\vspace{-1em}
\end{figure*}

\subsection{Decoders} \label{sec:decoders}
To reconstruct the masked-out tokens from the visible tokens, we use a separate decoder for each task.
The input to each decoder is the full set of visible tokens from the respective task it is reconstructing.
As in MAE~\cite{He2021MAE}, these visible tokens are decoded jointly with a set of mask tokens, which serve as \textit{placeholders} for the decoders to write the reconstructed patches (as shown in Fig.~\ref{fig:method_figure}).
To integrate information from the encoded tokens of other modalities, we add a single cross-attention layer in each decoder using these tokens as queries and all the encoded tokens as keys / values. Sine-cosine positional embeddings and learned modality embeddings are added to the tokens before this step.
This is then followed by a small MLP and Transformer blocks.
Following MAE, we compute the losses only on the masked tokens. 

As each task requires its own decoder, the computational cost of decoders scales linearly with the number of tasks. To keep pre-training efficient, we use shallow decoders (a single cross-attention layer and MLP, followed by two Transformer blocks) with a low dimensionality (256 dimensional).
Compared to the encoder, these decoders add little to the overall computational cost, and as He et al.~\cite{He2021MAE} show, they perform similarly to deeper decoders on ImageNet-1K fine-tuning.

\subsection{Multi-modal masking strategies} \label{sec:multimodal_masking}
For masked autoencoding to work well, a large percentage of tokens needs to be masked-out. He et al.~\cite{He2021MAE} showed that the choice of mask sampling strategy can have a large impact on transfer performance. More specifically for \mmae and generally learning multi-task representations, masking across different modalities ensures the model develops predictive coding across different modalities besides different spatial patches.
For efficiency and simplicity, we choose a constant number of visible tokens for all our experiments, which we fix at 98.
This corresponds to 1/6 of all tokens when using three modalities of dimensions 224$\times$224 pixels and a patch size of 16$\times$16.
Adapting the MAE mask sampling strategy by selecting the visible tokens uniformly from all tokens would result in most modalities being represented to similar degrees.
Cases where one or more modalities have very few or no samples would be very rare.
We propose a multi-modal token sampling strategy that allows for a more diverse sampling approach.
It can be broken down into two steps: First, selecting the number of tokens per modality, and second, randomly sampling the set of tokens for each modality.

\smallskip
\noindent\textbf{Number of tokens per modality.}
We select the proportion of tokens per modality $\lambda$ by sampling from a symmetric Dirichlet distribution $(\lambda_\text{\rgb}, \lambda_\text{\depth}, \lambda_\text{\semseg}) \sim \text{Dir}(\alpha)$, where $\lambda_\text{\rgb} + \lambda_\text{\depth} + \lambda_\text{\semseg} = 1, \lambda \geq 0$. 
The sampling is controlled by the concentration parameter $\alpha > 0$.
When $\alpha=1$, the symmetric Dirichlet distribution is equivalent to a uniform distribution over the simplex (i.e., it is uniform over all points in its support).
Smaller values ($\alpha << 1$) result in a sampling behavior where most of the tokens will be sampled from a single modality, while larger values ($\alpha >> 1$) result in an increasingly similar number of tokens to be sampled from each modality.
As a design decision, we do not bias the sampling towards certain modalities (as we use a symmetric Dirichlet), since we want to be agnostic to the choice of downstream input modalities and tasks that users might want to consider.
For simplicity and better representation of any possible sampled mask, we use a concentration parameter $\alpha=1$ for all of our experiments. Random masks sampled using $\alpha=1$ are shown in Figure~\ref{fig:dirichlet_samples}, and an ablation on the choice of concentration parameter is given in the Appendix (Sec. \ref{sec:mask_sampling}).

\smallskip
\noindent\textbf{Sampling tokens.}
From each modality, we sample the number of tokens, as specified by the above Dirichlet sampling step, uniformly at random without replacement.
Uniform sampling has been shown to work well for masked autoencoders, compared to less random alternatives~\cite{He2021MAE}.

\subsection{Pseudo labeled multi-task training dataset} \label{sec:pseudo_labeling}
We pre-train \mmae with three tasks that we pseudo label on ImageNet-1K~\cite{Deng2009ImageNet}. 
Pseudo labeling has the advantage that we do not need a large multi-task dataset with aligned task images.
Instead, having access to a good set of pre-trained neural networks for the tasks we want to train on can be effective. 
Pseudo labeling scales to RGB datasets of arbitrary size and is a one-time pre-processing step. 
Compared to the cost of training, this step is computationally cheap and fast if parallelized.

Taskonomy~\cite{Zamir2018Taskonomy} demonstrated computationally that common vision tasks cluster into three main categories, namely low-level, geometric, and semantic tasks.
To have a coverage over such a space of vision tasks, we choose one representative task from each of these three clusters.
We note that except for object detection and classification, these are the same pseudo labeled tasks that are used in MuST~\cite{Ghiasi2021MuST}.
In the following, we will describe them in more detail.

\smallskip
\noindent\textbf{RGB and per-patch standardized RGB.} 
We use RGB images due to their abundance and since RGB-only masked autoencoding is shown to be a powerful pre-training task.
He et al.~\cite{He2021MAE} study both predicting standard RGB patches, as well as per-patch standardized RGB patches.
They find that predicting standardized patches slightly improves transfer performance.
Since \mmae is naturally a multi-task model, we add both versions as separate decoder heads to get the representational benefits of predicting standardized patches, and to get a version that we can visualize better.
Note that we only add the per-patch standardized version as an output task, and not as an input modality.
For both RGB versions, we follow MAE and compute the MSE loss between the ground truth and predicted pixels.
In the rest of the paper, we will refer to the RGB and per-patch standardized RGB output tasks simply as RGB.

\smallskip
\noindent\textbf{Scene depth.} 
Depth is a key task informative about scene geometry.
As with RGB, but unlike semantic segmentation, sensors exist to capture this modality, making it possible to use depth as an optional extra input for downstream tasks.
To pseudo label depth, we use a DPT-Hybrid~\cite{Ranftl2021DPT} that was trained on Omnidata~\cite{Eftekhar2021Omnidata}.
Since monocular depth estimation is an inherently ill-posed task due to scale and shift ambiguity, we standardize the depth values in a robust way by ignoring the top and bottom 10\% of values~\cite{Yin2021LearningTR}. 
In addition, using standardized depth values as inputs allows us to use other depth images that might have different depth ranges and scales, without needing to match them to the Omnidata depth parameterization.
We use the L1 loss for depth.

\smallskip
\noindent\textbf{Semantic segmentation.}
Lastly, we use a Mask2Former~\cite{Cheng2021Mask2Former} with a Swin-S~\cite{liu2021swin} backbone trained on COCO~\cite{Lin2014COCO} to pseudo label semantic segmentation maps on ImageNet.
For that, we extract 133 semantic classes by taking the argmax of the network predictions.
Unlike RGB and depth, the main purpose of this task is to improve performance on downstream tasks, rather than using it as an input modality (though we show results using pseudo labeled semantic inputs in Table~\ref{tab:pseudo_transfers}).
Since we use a network that was pre-trained on COCO, we do not evaluate semantic segmentation transfers on that dataset.
For this task, we use the cross-entropy loss.

\subsection{Pre-training details} \label{sec:pretraining_details}
All our \mmae experiments use a \emph{ViT-B}~\cite{Dosovitskiy2021ViT} with a patch size of 16$\times$16 pixels.
We pre-train the models for either 400 epochs (only for transfer ablation study in Sec.~\ref{sec:task_choices}) or 1600 epochs (for best results and to be comparable to the MAE baseline) on 1.28M ImageNet images.
We use the AdamW~\cite{Loshchilov2019AdamW} optimizer with base learning rate 1e-4 and weight decay 0.05.
We warm up training for 40 epochs, starting from learning rate 1e-6, and decay it to 0 over the course of training using cosine decay~\cite{loshchilov2016sgdr}.
We set the batch size to a total of 2048 and train the models using 8 A100 GPUs with automatic mixed precision enabled.
Our data augmentations are straightforward.
We randomly crop the images, setting the random scale between 0.2 and 1.0 and the random aspect ratio between 0.75 and 1.33, after which we resize the crops to 224$\times$224 pixels and apply a random horizontal flip with probability 0.5. Additional pre-training details can be found in the Appendix (Sec. \ref{sec:pretraining_supp}).

\section{Experiments}
Optimizing the pre-training objective of \mmae is successful as apparent in the various results shown in the main paper, the appendix, and the interactive visualizations shown on \href{https://multimae.epfl.ch}{our website}.
In this section we provide a transfer study to measure the effectiveness of \mmae pre-training compared to relevant baselines.
This section is organized in the following manner:
After introducing the downstream tasks and datasets (Sec.~\ref{sec:transfer_tasks}), we show transfer results for the case where the only available input modality is RGB (Sec.~\ref{sec:rgb_transfers}).
Then, we show that \mmae can significantly improve downstream performance if other modalities like depth are either available as ground truth (sensor), or can be cheaply pseudo labeled (Sec.~\ref{sec:multimodal_transfers}).
We follow up with an ablation on the influence of pre-training tasks on the downstream performance (Sec.~\ref{sec:task_choices}), and finally we visually demonstrate that \mmae integrates and exchanges information across modalities (Sec.~\ref{sec:xmodal}).

\subsection{Transfer tasks and datasets} \label{sec:transfer_tasks}
We perform downstream transfers on a variety of semantic and dense regression tasks.
For all transfers, we replace the pre-trained decoders by randomly initialized task-specific heads, and train them along with the pre-trained encoder.
In the following, we give an overview over all tasks and datasets used in our transfer experiments.
Exact training details are presented in the Appendix (Sec. \ref{sec:transfer_supp}).

\smallskip
\noindent\textbf{Classification.} 
We evaluate our models and baselines by fine-tuning them on the supervised ImageNet-1K~\cite{Deng2009ImageNet} 1000-way object classification task.
We fine-tune our models for 100 epochs on the entire ImageNet-1K train split (1.28M images) and report the top-1 validation accuracy.

\smallskip
\noindent\textbf{Semantic segmentation.} 
We further evaluate our models on semantic segmentation tasks on the ADE20K~\cite{Zhou2017ADE20K} (20'210 training images and 150 classes), NYUv2~\cite{Silberman2012NYUv2} (795 training images and 40 classes), and Hypersim~\cite{Roberts2021Hypersim} (51'674 training images and 40 classes) datasets.
NYUv2 and Hypersim contain ground-truth depth maps that allow us to evaluate semantic segmentation with RGB and depth as input modalities.
For all datasets, we report the mean intersection over union (mIoU) metric.
On ADE20K and Hypersim, we report it on the validation split, while on NYUv2, we show the test set mIoU.

\smallskip
\noindent\textbf{Dense regression tasks.}
Finally, we study how our models transfer to geometric tasks, such as surface normals, depth and reshading, as well as tasks extracted from RGB images, such as keypoint or edge detection.
For depth estimation, we use NYUv2 (795 training and 655 test images), while for all other tasks we train transfers on a subset of the Taskonomy dataset~\cite{Zamir2018Taskonomy} (800 training images).
As performance metrics, we report $\delta_1$ on the NYUv2 test set, showing the percentage of pixels $p$ with error $\max\{\frac{\hat{y}_p}{y_p}, \frac{{y}_p}{\hat{y}_p}\}$ less than 1.25 \cite{doersch2017multi}, while on Taskonomy we report L1 losses on the tiny-split test set.

In the tables, classification, semantic segmentation, and depth estimation are denoted by (C), (S), and (D), respectively. 

\subsection{Transfers with RGB-only} \label{sec:rgb_transfers}
In this section, we show our transfer results when fine-tuning using only the RGB modality as input.

\smallskip
\noindent\textbf{Baselines.} For this setting, we compare \mmae with various \emph{ViT-B} models, namely \emph{DeiT}~\cite{Touvron2021DeiT} (without distillation) representing an ImageNet-supervised baseline, \emph{MoCo-v3}~\cite{Chen2021MoCoV3}, 
\emph{DINO}~\cite{Caron2021DINO}, and \emph{MAE}~\cite{He2021MAE}. All these models are pre-trained on ImageNet-1K.
We use the official weights for DeiT, MoCo-v3, and DINO, and reproduce MAE using the official PyTorch~\cite{Paszke2019PyTorch} codebase following the setting specified in \cite{He2021MAE} (i.e., decoder of depth 8 and width 512, per-patch standardized pixel loss, 1600 pre-training epochs, 75\% mask ratio). In the Appendix (Sec. \ref{sec:mae_diff_supp}), we compare the transfer performance of this MAE model to one with a shallower and narrower decoder (depth 2 and width 256), closer to the one used for \mmae.

We report the results in Table~\ref{tab:rgb_transfers}. We find that \mmae performs best on all tasks, matching MAE's performance on ImageNet-1K classification and ADE20K semantic segmentation, and outperforming it on all other tasks and datasets. These results show the effectiveness of \mmae as a pre-training strategy: it retains the benefits of MAE when RGB is the only fine-tuning modality but can also accept other modalities, as shown next.

\begin{table*}[t]
    \begin{minipage}[t]{.49\linewidth}
    \centering
    \resizebox{1.0\columnwidth}{!}{%
        \begin{tabular}{@{}lccccc@{}}
        \toprule
        Method & \small IN-1K (C) & \small ADE20K (S) & \small Hypersim (S) & \small NYUv2 (S) & \small NYUv2 (D) \\ \midrule
        Supervised \cite{Touvron2021DeiT}  & 81.8 & 45.8 & 33.9 & 50.1 & 80.7 \\
        DINO \cite{Caron2021DINO} & 83.1 & 44.6 & 32.5 & 47.9 & 81.3 \\
        MoCo-v3 \cite{Chen2021MoCoV3} & 82.8 & 43.7 & 31.7 & 46.6 & 80.9 \\
        
        MAE\cite{He2021MAE} & \textbf{83.3} & \textbf{46.2} & \underline{36.5} & \underline{50.8} & \underline{85.1} \\
        \midrule
        \mmae & \textbf{83.3} & \textbf{46.2} &  \textbf{37.0} & \textbf{52.0} & \textbf{86.4} \\ 
        \bottomrule
        \end{tabular}
        }
        \caption{ \textbf{Fine-tuning with \rgb-only.}
        We report the top-1 accuracy ($\uparrow$) on ImageNet-1K (IN-1K)~\cite{Deng2009ImageNet} classification (C), mIoU ($\uparrow$) on ADE20K~\cite{Zhou2017ADE20K} , Hypersim ~\cite{Roberts2021Hypersim} , and NYUv2~\cite{Silberman2012NYUv2}  semantic segmentation (S), as well as $\delta_1$ accuracy ($\uparrow$) on NYUv2 depth (D). Text in \textbf{bold} and \underline{underline} indicates the first and second-best results, respectively.
        All methods are pre-trained on ImageNet-1K (with pseudo labels for \mmae).
        }
    \label{tab:rgb_transfers}
    \end{minipage}
    \hfill
    \begin{minipage}[t]{.49\linewidth}
    \centering
        \resizebox{1.0\columnwidth}{!}{%
    \begin{tabular}{@{}lcccccccc@{}}
    \toprule
    && \multicolumn{3}{c}{Hypersim (S)} && \multicolumn{3}{c}{NYUv2 (S)} \\
    \cmidrule{2-5} \cmidrule{7-9}
    Method      && \rgb             & \depth            & \rgbd            && \rgb           & \depth           & \rgbd           \\ \midrule
    MAE && 36.5 &  \gc{32.5} & \gc{36.9} && 50.8 & \gc{23.4} & \gc{49.3} \\
    \mmae  && \textbf{37.0}             & \textbf{38.5}              & \textbf{47.6}             && \textbf{52.0}           & \textbf{41.4}             & \textbf{56.0 }           \\ \bottomrule
    \end{tabular}
    }
    \caption{\textbf{Fine-tuning with \rgb and ground truth \cdepth{depth}.}
    We report semantic segmentation transfer results from combinations of \rgb and \cdepth{depth}, measured in mIoU ($\uparrow$). \mmae can effectively leverage additional modalities such as \cdepth{depth}, while MAE cannot.
     Text in \gc{gray} indicates a modality that the model was not pre-trained on.
    }
    \label{tab:depth_transfers}
    \end{minipage}
\end{table*}

\begin{table*}[t]
    \centering
    \resizebox{\textwidth}{!}{%
    \begin{tabular}{@{}lcccccccccccccccccc@{}}
    \toprule
           && \multicolumn{5}{c}{ADE20K (S)} && \multicolumn{5}{c}{Hypersim (S)} && \multicolumn{5}{c}{NYUv2 (S)} \\
           \cmidrule{2-7} \cmidrule{9-13} \cmidrule{15-19}
           Method && \small \rgb    & \small \pdepth   & \small \rgbpd          & \small \rgbps & \small \rgbpdps  && \small \rgb     & \small \pdepth     & \small \rgbpd         & \small \rgbps & \small \rgbpdps     && \small \rgb    & \small \pdepth    & \small \rgbpd           & \small \rgbps & \small \rgbpdps   \\ \midrule
    MAE && \textbf{46.2} & \gc{20.0} & \gc{46.3} & \textbf{\gc{46.2}} & \gc{46.3} && 36.5 & \gc{21.0} & \gc{36.9} & \gc{37.7} & \gc{37.3} && 50.1 & \gc{23.8} & \gc{49.1} & \gc{50.1} & \gc{49.3} \\
    \mmae  && \textbf{46.2}   & \textbf{34.4}      & \textbf{46.8}   & 45.7 & \textbf{47.1}   &&  \textbf{37.0}     &  \textbf{30.6}        & \textbf{37.9}     & \textbf{38.4}  & \textbf{40.1}     && \textbf{52.0}    & \textbf{39.9}       & \textbf{53.6}    & \textbf{53.5}  & \textbf{54.0}   \\
    \bottomrule
    \end{tabular}
    }
    \caption{\textbf{Fine-tuning with \rgb and pseudo labels.}
    Semantic segmentation transfer results using \emph{pseudo labeled} \cdepth{depth} and \csemseg{semantic segmentation maps}, measured in mIoU ($\uparrow$). \mmae benefits much more than MAE from pseudo labeled modalities as input. Text in \gc{gray} indicates a modality that the model was not pre-trained on.
    }
    \label{tab:pseudo_transfers}
    \vspace{-1.5em}
\end{table*}

\subsection{Transfers with multiple modalities} \label{sec:multimodal_transfers}
Since \mmae was pre-trained on RGB, depth, and semantic segmentation, it can optionally accept any of those modalities as input during transfer learning should they be available.
In this set of experiments, we study on three semantic segmentation downstream tasks how much \mmae can benefit from using additional modalities during transfer.
Often, ground truth depth maps are not available for a given downstream dataset and for that reason, we perform additional transfers using pseudo labeled depth.
As there are several datasets that do in fact contain aligned RGB and depth images (e.g., Hypersim, NYUv2, Taskonomy, etc.) and since sensors exist that can measure depth, we consider it as a more realistic input modality compared to semantic segmentation.
Since our model was trained with semantic segmentation as an input modality, we perform additional experiments using pseudo labeled semantic segmentation maps as inputs.

All multi-modal transfers are performed by concatenating the projected patches of all modalities into a single sequence (i.e., no masking is performed here).
Using more than two modalities during transfer quickly becomes computationally expensive, since without masking, our method now scales with the full number of modalities and tokens.
For performing multi-modal transfers with the standard MAE, we train a new input projection for the additional modalities while fine-tuning.
Further training details can be found in the Appendix (Sec. \ref{sec:transfer_supp}).

\smallskip
\noindent\textbf{Transfers using sensory depth.} First, we consider that we have access to an aligned RGB-D dataset, like NYUv2 or Hypersim. We treat depth in the exact same way as during pre-training, i.e., pre-process it by standardizing it in a robust manner~\cite{Yin2021LearningTR}.
Because ground-truth depth maps might contain invalid measurements, we further set all these masked-out values to 0.

Table~\ref{tab:depth_transfers} shows RGB-D transfer results on Hypersim and NYUv2.
Compared to the RGB-only results in Table~\ref{tab:rgb_transfers}, we see a substantial increase in performance when ground truth depth is available for \mmae.
The standard MAE on the other hand is not able to sufficiently make use of the additional depth, since it was only trained on RGB images.
We observe a similar story when evaluating transfers from depth-only, in that \mmae works well, even when no RGB information is available, while MAE does not.
On Hypersim, \mmae depth-only transfer is even able to surpass \mmae RGB-only transfer, and, as expected, RGB-D works better than either RGB or depth alone.

\smallskip
\noindent\textbf{Transfers with pseudo labels.}
In case ground truth modalities are not available, we can pseudo label them in the same way we did for pre-training. To pseudo label depth, we use the same Omnidata DPT-Hybrid model that we used for pre-training on both ADE20K and NYUv2. On Hypersim, we use a MiDaS~\cite{Ranftl2022MiDaS} DPT-Hybrid, since the Omnidata depth model was partially trained on this dataset. For semantic segmentation pseudo labels, we use the same COCO Mask2Former model as in pre-training.

As shown in Table~\ref{tab:pseudo_transfers}, \mmae can use pseudo labeled depth or semantic segmentation to boost performance beyond the RGB-only setting, although the gain is smaller than using real depth.
Moreover, performance can further be improved by adding both of these pseudo labeled modalities to the input. This setting performs the best out of all settings involving pseudo labels.

\begin{table*}[t]
    \subfloat[
\textbf{Impact of additional modalities.} Transfer results of several \mmae models pre-trained on different input modalities / target tasks, compared against MAE (single-modal baseline). D2 = MAE pre-trained with a decoder of depth 2 and width 256, comparable in size to the decoders of \mmae 
\label{tab:ablation_same_in_out}
]{
    \begin{minipage}[t]{.48\linewidth}
    \centering
       \resizebox{1.0\columnwidth}{!}{%
        \begin{tabular}{@{}lcccc@{}}
        \toprule
        Method & \small IN-1K (C)  & \small NYUv2 (S)     & \small NYUv2 (D)     & \small Taskonomy (D) \\ \midrule
        MAE (D2)          & \underline{83.0} & 44.0          & 81.3          & 3.8           \\
        \rgbd           & 82.8          & 45.8          & 83.3          & \underline{2.1}           \\
        \rgbs           & \textbf{83.2} & \textbf{51.6} & \textbf{85.5} & 2.6           \\
        \rgbds          & \underline{83.0} & \underline{50.6} & \underline{85.4}          & \textbf{1.5}  \\ \bottomrule
        \end{tabular}
        }
    \end{minipage}
    }
    \hfill
    \subfloat[
\textbf{Comparison to non-masked pre-training}. We compare standard single-task and multi-task baselines pre-trained using \textit{non-masked} \rgb inputs against the \rgbds \mmae. The \crgb{RGB}$\rightarrow$\depthsemseg model is conceptually similar to MuST using depth and semantic segmentation as target tasks.
\label{tab:ablation_full_rgb}
]{
    \begin{minipage}[t]{.48\linewidth}
    \centering
     \resizebox{1.0\columnwidth}{!}{%
    \begin{tabular}{@{}lcccc@{}}
    \toprule
    Method                              & \small IN-1K (C)     & \small  NYUv2 (S)        & \small NYUv2 (D)        & \small Taskonomy (D)   \\ \midrule
    \crgb{RGB}$\rightarrow$\depth       & 82.7             & 44.0             & \textbf{87.1}    & \underline{1.6} \\
    \crgb{RGB}$\rightarrow$\semseg      & 82.5             & 46.8             & 82.9             & 4.0             \\
    \crgb{RGB}$\rightarrow$\depthsemseg & \underline{82.8} & \underline{48.6} & 84.6             & 2.9             \\ 
    \mmae                               & \textbf{83.0}    & \textbf{50.6}    & \underline{85.4} & \textbf{1.5}    \\ \bottomrule
    \end{tabular}
    }
    \end{minipage}
    } \vspace{-6pt}
    \caption{\textbf{Ablation experiments.} We study the impact of additional modalities in Table \ref{tab:ablation_same_in_out}, and compare \mmae to non-masked pre-training in Table \ref{tab:ablation_full_rgb}. All models are pre-trained for 400 epochs. We report the top-1 accuracy ($\uparrow$) on ImageNet-1K (IN-1K)~\cite{Deng2009ImageNet} classification (C), mIoU ($\uparrow$) on NYUv2~\cite{Silberman2012NYUv2} semantic segmentation (S), $\delta_1$ accuracy ($\uparrow$) on NYUv2 depth (D) and avg. rank $\downarrow$ on Taskonomy~\cite{Zamir2018Taskonomy}. While some specialized pre-trained models perform better at certain downstream tasks, they perform poorly at others. \mmae pre-trained with \rgb, \cdepth{depth} and \csemseg{semantic segmentation} is a more generalist model that does well at transferring to a range of downstream tasks.}
    \label{tab:ablations}
\end{table*}

\subsection{Influence of pre-training task choices and masking on transfer performance} \label{sec:task_choices}
How does the choice of \mmae pre-training tasks affect downstream transfer performance?
In this subsection, we aim to address this question by performing transfers from \mmae models that were pre-trained with \rgbd, \rgbs, or \rgbds.
We further compare \mmae against MAE, single-task, and multi-task baselines.

All experiments are performed on ViT-B models that were pre-trained for 400 epochs.
We transfer the pre-trained models to ImageNet, NYUv2 segmentation, as well as nine dense regression tasks on Taskonomy.
On Taskonomy, we report the ranking of different pre-trained models, averaged over all nine tasks.
Detailed per-task results on Taskonomy can be found in the Appendix (Sec. \ref{sec:taskonomy_supp}).

\begin{figure*}[t]
\centering
\includegraphics[width=0.9\textwidth]{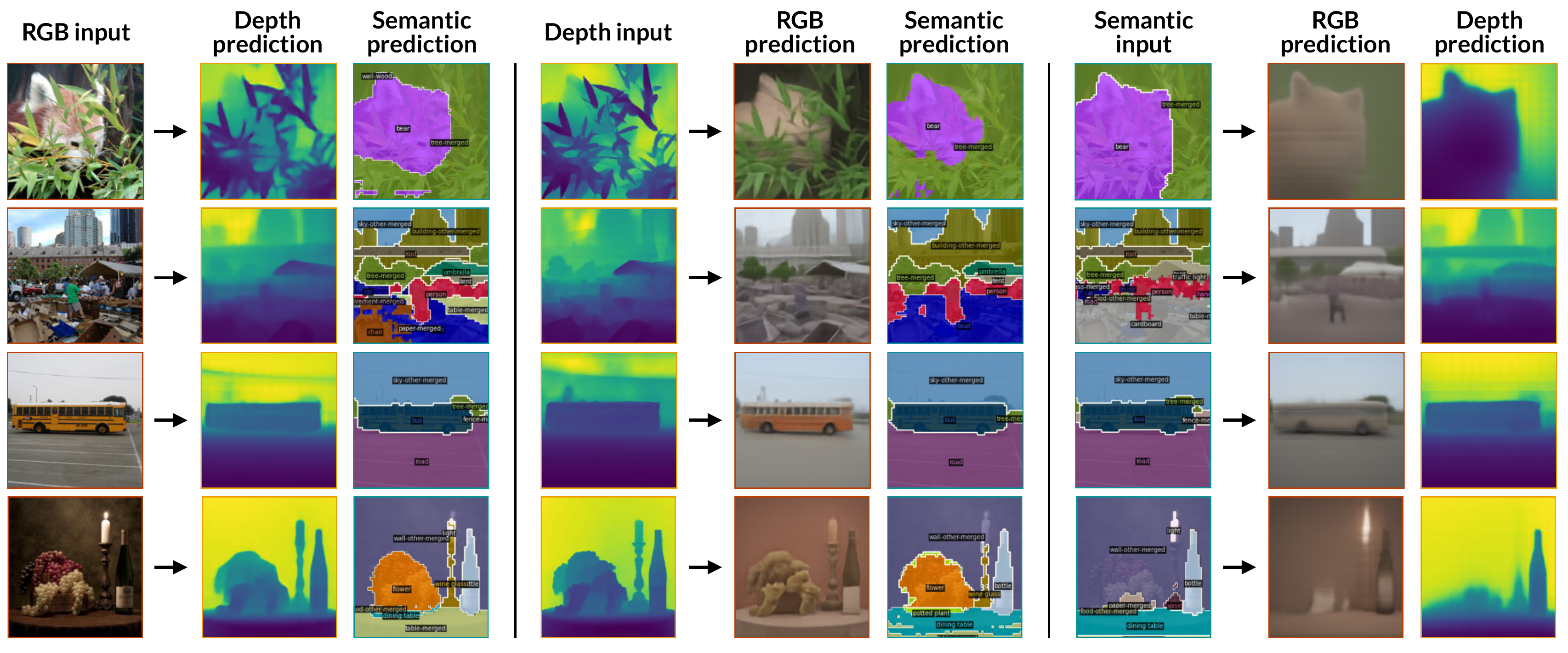}
\caption{\textbf{Single-modal predictions.}
We visualize \mmae cross-modal predictions on ImageNet-1K validation images. Only a single, full modality is used as input.
The predictions remain plausible despite the absence of input patches from other modalities.
}
\label{fig:one_to_two}
\centering
\vspace{-0.5em}
\end{figure*}

\smallskip
\noindent\textbf{Masked multi-modal pre-training.}
This experiment studies the influence that the choice of pre-training modalities has, when the input and output modalities are the same in \mmae pre-training.
The transfer results are displayed in Table~\ref{tab:ablation_same_in_out}.
The \rgbs model performs best on ImageNet classification and NYUv2 semantic segmentation, whereas the \rgbds model has the best average rank on Taskonomy.
The slight increase in performance of \rgbs on ImageNet and semantic segmentation compared to \rgbds comes at the cost of reduced flexibility, as models that were not pre-trained on depth can not as easily and effectively use it to boost performance (see Sec.~\ref{sec:multimodal_transfers}).

\smallskip
\noindent\textbf{Comparison to non-masked pre-training.}
We further compare \mmae against standard single-task and multi-task baselines, that were pre-trained with RGB as the only input modality and without applying any masking.
Since we train on pseudo labels, the \crgb{RGB}$\rightarrow$\depthsemseg multi-task model is conceptually similar to a MuST~\cite{Ghiasi2021MuST} model using depth and semantic segmentation targets.
The transfer results are detailed in Table~\ref{tab:ablation_full_rgb}.
On nearly all categories, \mmae outperforms the supervised baselines.

To summarize, the results in this section show that using all modalities to pre-train a \mmae results in a more generalist model that does well at transferring to a range of downstream tasks.
We find that there are some \textit{specialized} pre-trained models that perform better at certain downstream tasks (e.g., models pre-trained with depth perform better at transferring to geometric tasks), but they will perform poorly at others.
This is supported by previous findings~\cite{Zamir2018Taskonomy, Sax2018MidLevel, Mensink2021Factors} showing that there is usually no single visual pre-training task that transfers well to any arbitrary other task, and instead, a set is required.

\subsection{Cross-modal exchange of information} \label{sec:xmodal}
In this section, we explore visually how \mmae predicts the three pre-training tasks by changing the inputs it receives.
Figures~\ref{fig:random_samples} and \ref{fig:dirichlet_samples} already showcased how \mmae is able to reconstruct images from various randomly sampled input patches.
Here, we will further show non-masked cross-modal predictions, and will also give examples on how \mmae predictions change when we change certain details about the inputs.

\begin{figure*}[t]
\centering
\includegraphics[width=\textwidth]{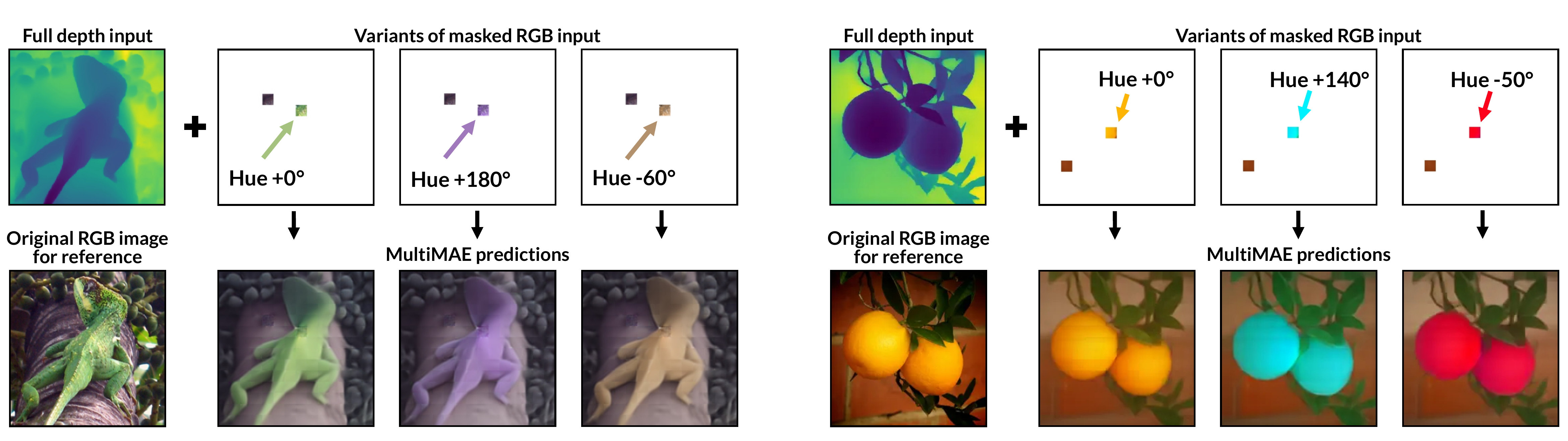}
\caption{\textbf{Demonstration of cross-modal interaction.} The input is the full depth, only two RGB patches, and \emph{no} semantic segmentation.
By editing the hue of a single input patch, the color of the lizard (left) and oranges (right) changes, while keeping the background constant. More interactive examples are available on \href{https://multimae.epfl.ch\#hue-change}{our website}.
}
\label{fig:task_interaction}
\centering
\vspace{-1em}
\end{figure*}

\begin{figure}[t]
\centering
\includegraphics[width=\columnwidth]{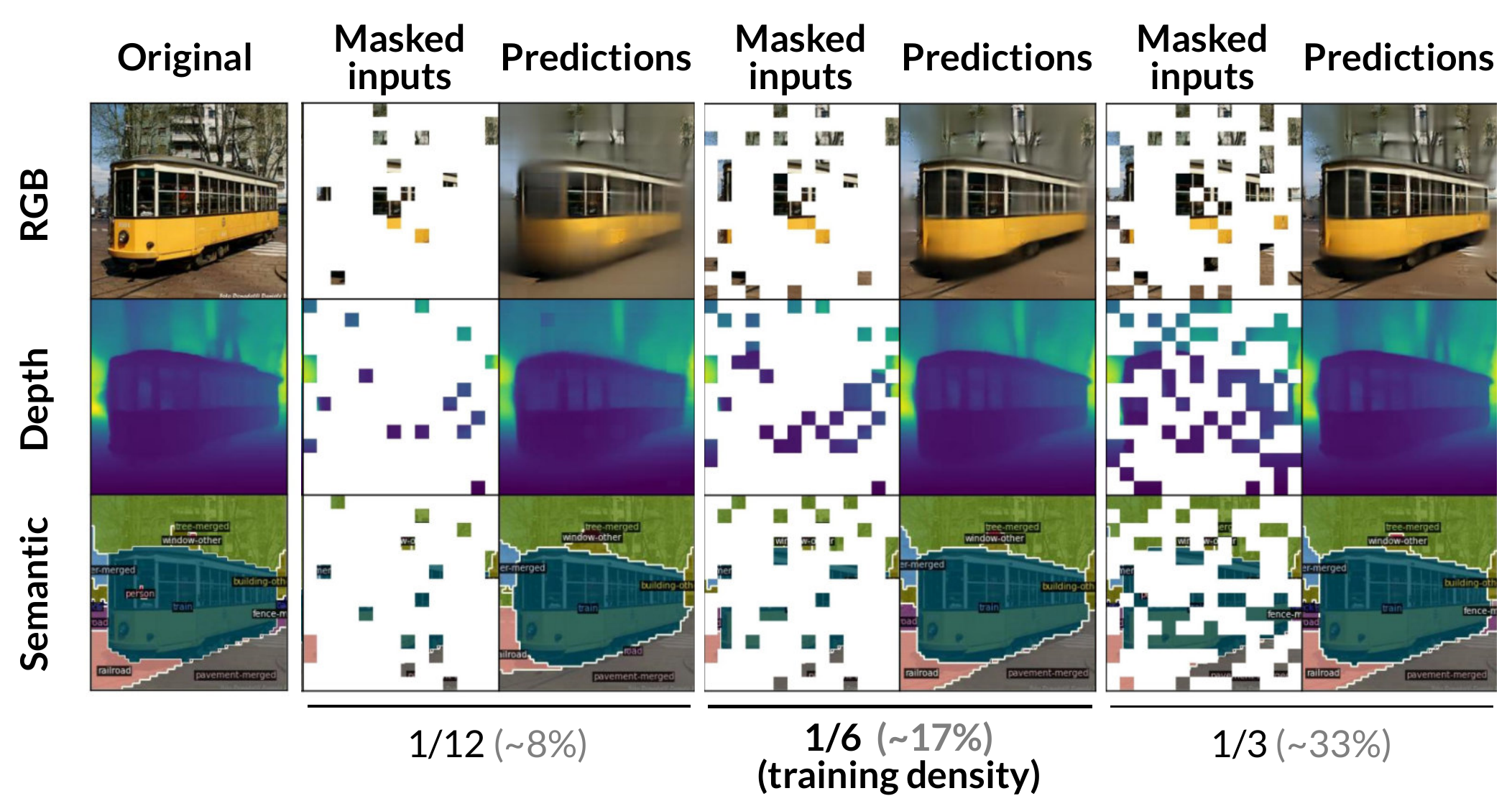}
\caption{
    \textbf{\mmae predictions for a varying number of visible patches.} The predictions are plausible even when given half the number of patches seen during pre-training, and the reconstruction quality improves as the number of visible patches increases. An interactive visualization is available on \href{https://multimae.epfl.ch/\#masking-percentage}{our website}.
}
\vspace{-1em}
\label{fig:masking_percentage}
\centering
\vspace{-0.5em}
\end{figure}
\smallskip
\noindent\textbf{Single-modal predictions.} 
Figure~\ref{fig:one_to_two} displays several examples of cross-modal prediction without any masking.
We show examples where, from one single modality, the two remaining ones are predicted. We note here that even though the number of patches we input to the model is 2$\times$ higher than what was seen during training, the model still predicts very reasonable results despite the distribution shift. The robustness of the model with respect to masking ratios far from the training mask ratio is also apparent in Figure~\ref{fig:masking_percentage}.

\smallskip
\noindent\textbf{Demonstration of cross-modal interaction.}
We demonstrate in Figure~\ref{fig:task_interaction} how \mmae predicts completely different but plausible RGB images when given a full depth image and three edited versions of the same two RGB input patches (no semantic segmentation maps are given as inputs). We keep one RGB patch the same, while changing the hue of another patch (part of a lizard for the first image, part of an orange for the second).
We can see how \mmae recovers all the details in the image from the full depth input, but paints the entire lizard / oranges in the colors given in the modified patch.
All the while, the background does not change. This suggests an intriguingly good representation is learned by the model as it extends the colors to the right segments without any segmentation provided in the input. More interactive examples can be seen on \href{https://multimae.epfl.ch\#hue-change}{our website}.

\section{Discussion}
We presented Multi-modal Multi-task Masked Autoencoders (\mmae), an effective and simple pre-training strategy for Vision Transformers.
\mmae encodes a small random subset of visible tokens from multiple modalities and is trained to reconstruct the missing ones.
By encoding only a fixed number of non-masked tokens, we can keep the bulk of the computation in the Transformer encoder constant, while only the shallow task-specific decoders scale with the number of tasks. Masking (across image patches and input modalities) ensures the network learns to perform predictive coding across different modalities, besides across different spatial patches.
The experiments showed intriguing capabilities of \mmae at cross-modal coding and demonstrated this pre-training strategy can result in notable gains in transfer performance when additional input modalities are optionally available, either as ground truth or pseudo labels.

In the following, we briefly discuss some limitations to our approach and present exciting future directions:

\smallskip
\noindent\textbf{Scaling pre-training modalities.} 
We pre-trained \mmae on a set of three visual modalities, chosen to cover a large fraction of common vision problems based on prior studies~\cite{Zamir2018Taskonomy}.
It is, however, conceivable that our method can benefit from a rather straightforward inclusion of a more diverse set of modalities and tasks, such as videos, text, bounding boxes, sparse depth, feature maps, and more.
In addition to providing more ways to use optional modalities as inputs, scaling up the number of pre-training modalities could have further transfer benefits by covering a larger space of useful vision problems and enabling more complex cross-modal predictive coding.

\smallskip
\noindent\textbf{Scaling pre-training datasets.}
For pragmatic reasons and enabling comparison with prior works, we trained all of our models on pseudo labeled ImageNet-1K, but there is no reason to limit ourselves to a (classification) dataset of this size.
Since we use pseudo labels, any dataset that is used for RGB-only self-supervised learning can be considered for training \mmae.
Our method further benefits from any future improvements in model architectures, training strategy and supervised datasets that can be used to improve the quality of pseudo labels.

\smallskip
\noindent\textbf{Probabilistic or generative modeling.}
Similar to standard autoencoders and MAE~\cite{He2021MAE}, we simply compute a pixel-wise L1 or L2 loss on the reconstructed tokens.
It is unsurprising then that ambiguous masked regions are often predicted in a blurry manner due to the inherent ambiguity in the problem when multiple outputs are plausible.
While He et al.~\cite{He2021MAE} showed that improving the visual fidelity of MAE predictions might not necessarily result in better representations for downstream learning, it is conceivable that modeling the multi-modal output distribution may learn better representations.

\smallskip
\noindent\textbf{Masking strategies.}
Lastly, we used a simple approach of sampling random tokens from each modality in an unbiased way.
While this worked well for \mmae training, it does not have to be the optimal choice for learning a transferable representation. It will be an interesting direction to explore biasing the masking towards certain modalities and/or spatial locations.

\section*{Acknowledgments}
We thank Stefan Stepanovic and Alexander Sax for their help and insightful discussions.


{\small
\bibliographystyle{ieee_fullname}
\bibliography{references}
}
\clearpage

\title{\mmae: Multi-modal Multi-task Masked Autoencoders \\ \textit{Appendix} \vspace{-4mm}}

\author{Roman Bachmann\footnotemark[1] \quad\quad David Mizrahi\footnotemark[1] \quad\quad Andrei Atanov \quad\quad Amir Zamir\\
Swiss Federal Institute of Technology Lausanne (EPFL) \vspace{3mm}\\
\url{https://multimae.epfl.ch}
}

\maketitle

\footnotetext[1]{Equal contribution.}

\appendix

\addcontentsline{toc}{section}{} 
\part{} 
\vspace{-3em}
\parttoc 

\section{Additional pre-training implementation details} \label{sec:pretraining_supp}

We report the default pre-training setting in Table~\ref{tab:pretraining-setting}. The learning rate follows the linear scaling rule \cite{goyal2017accurate}: $\textit{lr} = \textit{base\_lr}\times \text{batchsize} / 256$. 
The number of non-masked tokens given to the encoder is set to 49 when using a single input modality (mask ratio of 3/4), and 98 when using 2 or 3 modalities (mask ratio of 3/4 and 5/6, respectively). 
Furthermore, given that the semantic segmentation map consists of 64-dimensional class embeddings, naively projecting each patch to a token is computationally expensive (when flattened, each patch would have a dimension of 16384 and the projection layer would have approx. 12M parameters). To make this projection efficient while keeping the number of segmentation patches constant, we downsample the semantic segmentation input by a factor of 4 and use patches of size 4$\times$4.

\begin{figure}[htpb]
\centering
\includegraphics[width=1.0\columnwidth]{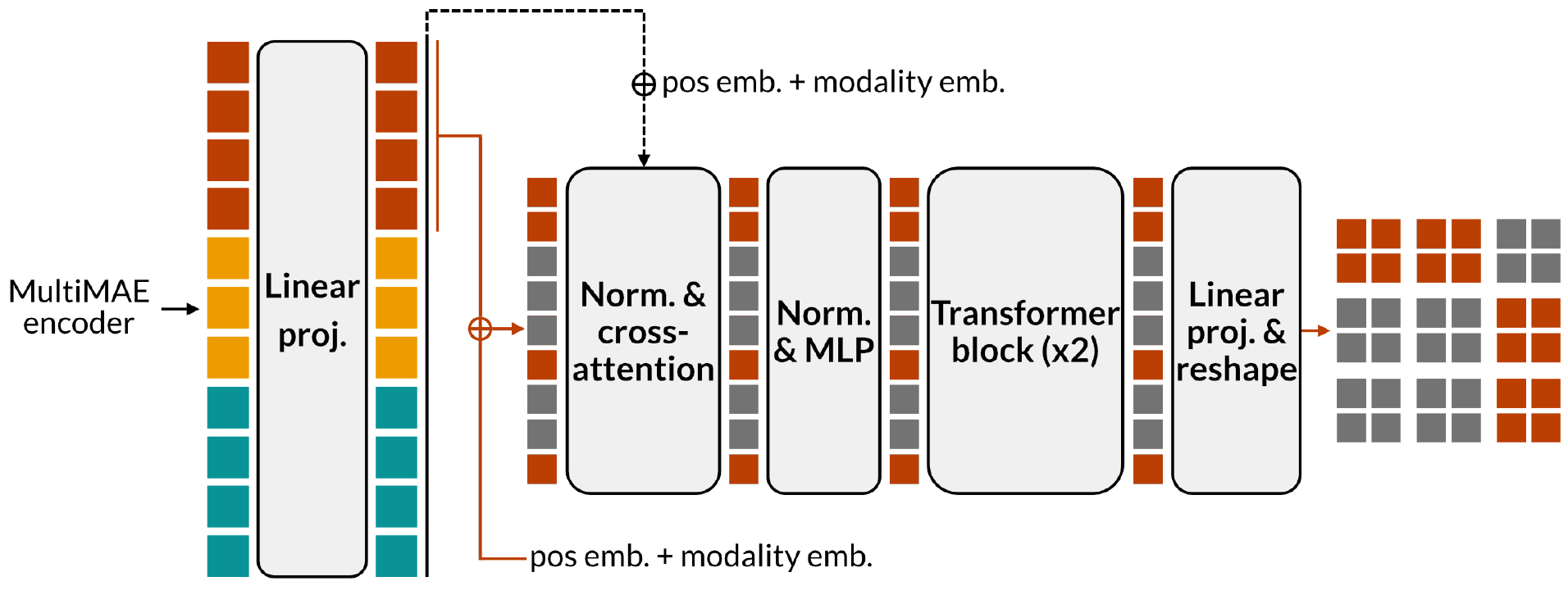}
\caption{
    \textbf{\mmae decoders}: 
    Tokens from the \mmae encoder (see Fig. \ref{fig:method_figure}) are first linearly projected to the decoder dimension, after which positional and modality-specific embeddings are added.
    A cross-attention step integrates information from tokens of other modalities before applying an MLP and two Transformer blocks.
    Finally, each token is projected and reshaped to form an image.
    In this illustration, each token expands into four pixels.
}
\label{fig:decoder}
\centering
\vspace{-1em}
\end{figure}

\smallskip
\noindent\textbf{MultiMAE decoder.} 
We illustrate the \mmae decoder in Fig~\ref{fig:decoder}. Following MAE~\cite{He2021MAE}, each decoder has a linear projection layer to adapt the outputs from the encoder to the decoder dimension. After this linear projection, we add both sine-cosine positional embeddings and learned modality embeddings to the decoder inputs. This is then followed by a cross-attention layer, a MLP, and two Transformer blocks.

\section{Transfer implementation details} \label{sec:transfer_supp}

\subsection{ImageNet classification fine-tuning setting}

For ImageNet-1K~\cite{Deng2009ImageNet} classification, we follow the end-to-end fine-tuning procedure from MAE~\cite{He2021MAE} and replace the decoders by an average pooling operation over all encoded tokens, followed by LayerNorm~\cite{ba2016layer} and a linear projection. The default setting is shown in Table~\ref{tab:classification-setting}.

\begin{table*}[htpb]
    \begin{minipage}[t]{.47\textwidth}
      \centering
      \begin{adjustbox}{max width=\textwidth}
        \begin{tabular}{l|l}
        \toprule
        Hyperparameters       & Value              \\ 
        \midrule
        Optimizer              & AdamW \cite{Loshchilov2019AdamW}                 \\
        Base learning rate \cite{goyal2017accurate}     & 1e-4                   \\
        Weight decay           & 0.05                   \\
        Adam $\beta$             & (0.9, 0.95)            \\
        Batch size             & 2048                   \\
        Learning rate sched. & Cosine decay \cite{loshchilov2016sgdr}           \\
        Training epochs & 400 or 1600 \\
        Warmup learning rate & 1e-6 \\
        Warmup epochs          & 40                     \\ 
        \midrule
        Non-masked tokens & 98\\
        Sampling $\alpha$       & 1.0                    \\
        Task weighting         & None (equal weights)   \\
        \midrule
        Input resolution                            & 224 $\times$ 224              \\
        Augmentation                              & RandomResizedCrop      \\

        \bottomrule
        \end{tabular}
        \end{adjustbox}
        \caption{\textbf{Default pre-training setting.} For ablations, the number of epochs is set to 400. For best results, the number of epochs is set to 1600.}
        \label{tab:pretraining-setting}
    \end{minipage}
    \hfill
    \begin{minipage}[t]{.47\textwidth}
      \centering
      \begin{adjustbox}{max width=\textwidth}
        \begin{tabular}{l|l}
        \toprule
        Hyperparameters        & Value                \\ 
        \midrule
        Optimizer              & AdamW \cite{Loshchilov2019AdamW}                 \\
        Base learning rate \cite{goyal2017accurate}    & 5e-4               \\
        Weight decay           & 0.05                   \\
        Adam $\beta$             & (0.9, 0.999)            \\
        Layer-wise lr decay\cite{clark2020electra} & 0.65 \\ 
        Batch size             & 1024                   \\
        Learning rate sched. & Cosine decay \cite{loshchilov2016sgdr}           \\
        Training epochs & 100 \\
        Warmup learning rate & 1e-6 \\
        Warmup epochs          & 5                     \\ 
        \midrule
        Input resolution                            & 224 $\times$ 224              \\
        Augmentation                              & RandAugment(9, 0.5)      \\
        Label smoothing                         & 0.1 \\
        Mixup \cite{zhang2017mixup} & 0.8 \\
        Cutmix \cite{yun2019cutmix} & 1.0 \\
        Drop path \cite{huang2016deep} & 0.1 \\ 
        \bottomrule
        \end{tabular}
        \end{adjustbox}
        \caption{\textbf{ImageNet-1K classification setting.} We follow the fine-tuning settings from MAE~\cite{He2021MAE}.
        }
        \label{tab:classification-setting}
    \end{minipage}
\end{table*}

\subsection{Semantic segmentation}

The typical approach \cite{Bao2021BEiT, He2021MAE} to fine-tuning Vision Transformers for semantic segmentation is not suited for multi-modal inputs in two aspects: 1) the segmentation head and 2) the evaluation procedure. We cover these two aspects next and propose a simplified fine-tuning setting for semantic segmentation to overcome these issues.

\smallskip
\noindent\textbf{Segmentation head.} The UPerNet~\cite{xiao2018unified} head used in BEiT~\cite{Bao2021BEiT} and MAE~\cite{He2021MAE} operates on a feature pyramid \cite{lin2017feature}. While a Vision Transformer operating only on RGB images can be modified to return hierarchical feature maps through the use of deconvolution layers on intermediate features \cite{zheng2021rethinking}, this procedure is not so simple when the input is multi-modal.
In contrast, segmentation heads that operate \emph{only} on the output tokens do not have this issue. One such head is the Segmenter~\cite{strudel2021segmenter}, for which tokens are passed through additional Transformer blocks, then reshaped into a feature map and upsampled to full resolution. 
However, the direct upsampling can result in inprecise segmentation maps and hurt performance. Instead, we propose using a simple segmentation head based on the ConvNeXt architecture~\cite{liu2022convnet}. First, we increase the dimensionality $D$ of the output tokens wih a linear projection, and then reshape the tokens to form a feature map of size $H/4 \times W/4 \times D/8$. We then apply 4 ConvNeXt blocks on this feature map before upsampling it to full resolution using bilinear interpolation.
We find that this simple ConvNeXt head outperforms Segmenter, as shown in Table~\ref{tab:semseg-head}. To adapt this head to multi-modal inputs, we can either select only the output tokens from a single modality (as information from other modalities gets passed to these tokens through self-attention) or concatenate tokens from different modalities. We find that both approaches perform comparably and select the former as it is slightly more efficient.

\begin{table}[t]
\centering
\resizebox{1.0\columnwidth}{!}{%
\begin{tabular}{@{}llccc@{}}
\toprule
Method & Head & ADE20K  & Hypersim & NYUv2  \\ \midrule
\mmae & Segmenter-Mask~\cite{strudel2021segmenter} &\textbf{46.3} & 36.0 &  49.0 \\
\mmae & ConvNeXt & 46.2 & \textbf{37.0} & \textbf{52.0}  \\
\bottomrule
\end{tabular}
}
\caption{\textbf{Comparison of semantic segmentation heads.} We report the mIoU ($\uparrow$) on ADE20K~\cite{Zhou2017ADE20K}, Hypersim~\cite{Roberts2021Hypersim} and NYUv2~\cite{Silberman2012NYUv2}. The proposed segmentation head based on the ConvNeXt~\cite{liu2022convnet} architecture performs on average slightly better than Segmenter~\cite{strudel2021segmenter}.}
\label{tab:semseg-head}
\vspace{-2em}
\end{table}

\smallskip
\noindent\textbf{Evaluation procedure.} Vision Transformers are commonly evaluated using the sliding window procedure from MMSegmentation~\cite{MMSegmentation_Contributors_OpenMMLab_Semantic_Segmentation_2020} (e.g., \cite{zheng2021rethinking, bao_beit_2021, zhou2021ibot}). This procedure involves first resizing the validation images so that the \emph{smallest side} matches the training resolution\footnote{In some implementations, the height is resized to the training resolution, which most often coincides with the smallest side.}, and then applying a sliding window over the resized image and averaging predictions across windows. However, this procedure is not suitable if the input modalities rely on statistics from the entire image (e.g., standardized depth) or do not have a 2D structure (e.g., object bounding boxes). Therefore, we use a simpler evaluation procedure inspired by \cite{li_benchmarking_2021}, which consists of resizing the image so that the \emph{largest side} matches the training resolution and padding the smallest side. As the evaluated images have a smaller resolution, this simple procedure results in slightly worse reported performance compared to sliding windows. However, it can be used regardless of the input modalities and thus allows for a more fair comparison of segmentation performance for different modalities.

\smallskip
\noindent\textbf{Training details.} The semantic segmentation transfer settings for all three segmentation datasets are shown in Table~\ref{tab:semseg-setting}. Following \cite{li_benchmarking_2021}, our main augmentation is large scale jittering (LSJ) \cite{ghiasi2021simple}. We also apply color jittering with the following parameters: \texttt{brightness=0.4, contrast=0.4, saturation=0.2, hue=0.1, p=0.5}.

\begin{table}[t]
\centering
\resizebox{1.0\columnwidth}{!}{%
\begin{tabular}{l|ccc}
\toprule
 Hyperparameters & ADE20K & Hypersim~ & NYUv2 \\
\toprule
Optimizer & \multicolumn{3}{c}{AdamW \cite{Loshchilov2019AdamW}} \\
Learning rate & \multicolumn{3}{c}{1e-4} \\
Layer-wise lr decay\cite{clark2020electra} & \multicolumn{3}{c}{0.75} \\
Weight decay & \multicolumn{3}{c}{0.05} \\
Adam $\beta$ & \multicolumn{3}{c}{(0.9, 0.999)} \\
Batch size & 16 & 16 & 8 \\
Learning rate sched. & \multicolumn{3}{c}{Cosine decay \cite{loshchilov2016sgdr}} \\
Training epochs & 64 & 25 & 200 \\
Warmup learning rate & \multicolumn{3}{c}{1e-6} \\
Warmup epochs & \multicolumn{3}{c}{1} \\
\midrule
Input resolution & 512 $\times$ 512 & 512 $\times$ 512 & 640 $\times$ 640 \\
Augmentation & \multicolumn{3}{c}{Large scale jittering (LSJ) \cite{ghiasi2021simple}} \\
Color jitter & \multicolumn{3}{c}{\cmark} \\
Drop path \cite{huang2016deep} & \multicolumn{3}{c}{0.1} \\
\bottomrule
\end{tabular}
}
\caption{
\textbf{Semantic segmentation fine-tuning settings }for ADE20K~\cite{Zhou2017ADE20K}, Hypersim~\cite{Roberts2021Hypersim} and NYUv2~\cite{Silberman2012NYUv2}.
}
\label{tab:semseg-setting}
\end{table}

\subsection{NYUv2 depth estimation}

For depth estimation on the NYUv2 dataset.~\cite{Silberman2012NYUv2}, we resize all images from 640~$\times$~480 to $341~\times~256$. During training, we randomly crop the images to 256~$\times$~256 and during testing, we take a central crop of size 256~$\times$~256.

We follow \cite{grill2020bootstrap, laina2016deeper} and apply color jittering with the following parameters: \texttt{brightness=0.1255, contrast=0.4, saturation=0.5, hue=0.2, p=0.5}.
We also randomly turn the image into gray-scale with probability $p=0.3$.

We use the DPT~\cite{Ranftl2021DPT} head to decode layers [3,6,9,12] of the ViT-B encoder into the dense depth map. For training, we use the reverse Huber loss \cite{laina2016deeper}.
Detailed transfer settings are given in Table~\ref{tab:regression_transfers}.
For evaluation, we measure the $\delta_1$ metric on the test set, showing the percentage of pixels $p$  with error $\max\{\frac{\hat{y}_p}{y_p}, \frac{{y}_p}{\hat{y}_p}\}$ less than 1.25.

\addtolength{\tabcolsep}{3pt}    
\begin{table}[htpb]
\centering
\resizebox{1.0\columnwidth}{!}{%
\begin{tabular}{@{}l|cc@{}}
\toprule
Hyperparameters & \multicolumn{1}{l}{NYUv2 depth} & \multicolumn{1}{l}{Taskonomy tasks} \\ \midrule
Optimizer & \multicolumn{2}{c}{AdamW~\cite{Loshchilov2019AdamW}} \\
Learning rate & 1e-4 & 3e-4 \\
Layer-wise lr decay \cite{clark2020electra} & \multicolumn{2}{c}{0.75} \\
Weight decay & 1e-4 & 5e-2 \\
Adam $\beta$ & \multicolumn{2}{c}{(0.9, 0.999)} \\
Batch size & 128 & 32 \\
Learning rate sched. & \multicolumn{2}{c}{Cosine decay~\cite{loshchilov2016sgdr}} \\
Training epochs & 2000 & 100 \\
Warmup learning rate & \multicolumn{2}{c}{1e-6} \\
Warmup epochs & 100 & 5 \\ \midrule
Input resolution & 256 $\times$ 256 & 384 $\times$ 384 \\
RandomCrop & \cmark & \xmark \\
Color jitter & \cmark & \xmark \\
Drop path \cite{huang2016deep} & \xmark & 0.1 \\ \bottomrule
\end{tabular}
}
\caption{
\textbf{Fine-tuning settings for NYUv2~\cite{Silberman2012NYUv2} depth estimation and eight Taskonomy~\cite{Zamir2018Taskonomy} 2D regression tasks}.
}
\label{tab:regression_transfers}
\end{table}
\addtolength{\tabcolsep}{-3pt}

\subsection{Taskonomy dense regression tasks} 
We transfer to the following eight dense regression tasks from the Taskonomy~\cite{Zamir2018Taskonomy} dataset:
\textit{Principal curvature, z-buffer depth, texture edges, occlusion edges, 2D keypoints, 3D keypoints, surface normals, and reshading.}
We train the transfers on a random subset of the Taskonomy-tiny split, selecting 800 training and 200 validation images.
The test evaluation is performed on the entire Taskonomy-tiny test split (54514 images), using the checkpoint with the lowest validation loss.

For training and testing, all images are resized to $384 \times 384$ and we perform no further augmentations.
As for NYUv2~\cite{Silberman2012NYUv2} depth estimation, we use the DPT~\cite{Ranftl2021DPT} head, accessing layers [3,6,9,12] from the ViT-B encoder.
All tasks are trained with an L1 loss.
Detailed transfer settings are given in Table~\ref{tab:regression_transfers}.

\section{Mask sampling strategies} \label{sec:mask_sampling}
\begin{figure}[t]
\centering
\includegraphics[width=1.0\columnwidth]{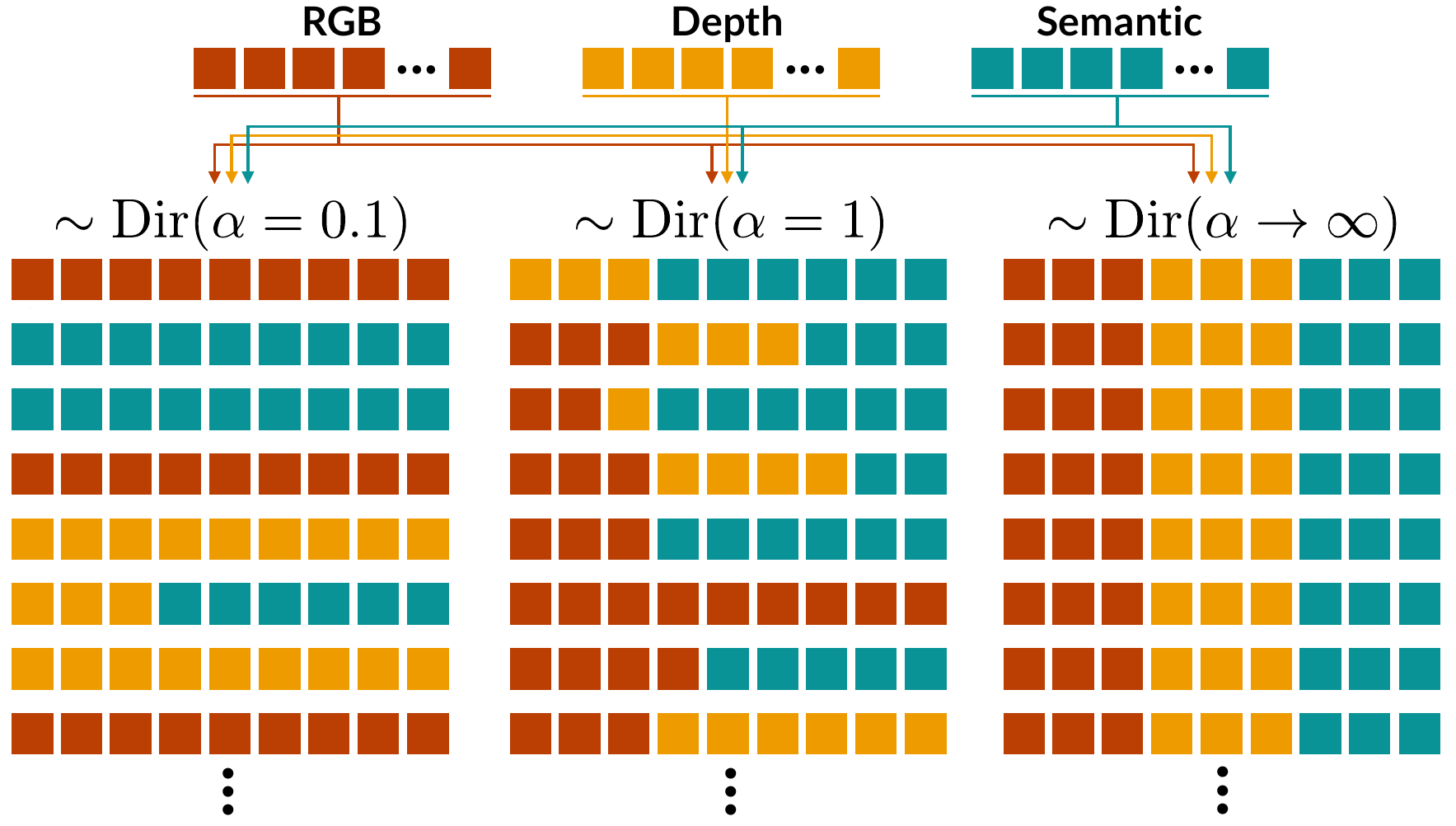}
\vspace{-1em}
\caption{
\textbf{Multi-modal mask sampling}:
We sample the proportion of tokens per modality using a symmetric Dirichlet distribution $\text{Dir}(\alpha)$ with concentration parameter $\alpha$.
We illustrate here the sampling behavior for different choices of $\alpha$ values when selecting nine tokens from three modalities. Each row represents one sample of tokens.
With small $\alpha$, most tokens will be sampled from single modalities, while large $\alpha$ values result in equal representation of each modality.
Setting $\alpha = 1$ is equivalent to sampling uniformly over the support and results in a more diverse sampling behavior.
}
\label{fig:dirichlet}
\centering
\vspace{-0.5em}
\end{figure}

We sample the number of non-masked tokens per modality using a Dirichlet distribution with concentration parameter $\alpha = 1$.
Figure~\ref{fig:dirichlet} illustrates the sampling behavior under different $\alpha$ values.
For simplicity, we picked $\alpha = 1$ for all our experiments in the main paper, which exposes the models to a large diversity of masks.
Samples using $\alpha = 1$ include cases where all tokens are sampled from a single modality (very low $\alpha$) and \mmae has to fully reconstruct the other two, cases where all modalities are equally represented (very high $\alpha$), and everything in between.

In Table~\ref{tab:alpha_ablation}, we show transfer results on ImageNet-1K~\cite{Deng2009ImageNet} classification and ADE20K~\cite{Zhou2017ADE20K} semantic segmentation using \mmae models trained with $\alpha \in \{0.2, 0.5, 1.0, \infty \}$.
By $\alpha = \infty$, we denote always sampling an equal number of tokens from each modality.
All models in this table were trained for 400 epochs and do not include the additional per-patch-standardized \rgb head (see Sec. \ref{sec:pseudo_labeling}).
Setting $\alpha = 1$ performs best on ADE20K, while being close second on ImageNet-1K behind $\alpha = \infty$.
Smaller values of $\alpha$ do not perform better on these two \rgb-only downstream tasks, even though during training they were exposed to more samples that contain tokens from only one modality.
Biasing the sampling towards modalities that will be used during transfer is an interesting future direction.

\begin{table}[t]
\centering
\begin{tabular}{@{}lcc@{}}
\toprule
$\alpha$ & ImageNet-1K~\cite{Deng2009ImageNet} & ADE20K~\cite{Zhou2017ADE20K} \\ \midrule
0.2 & 82.7 & 44.6 \\
0.5 & 82.5 & {\ul 44.8} \\
1.0 & {\ul 82.8} & \textbf{45.1} \\
$\infty$ & \textbf{82.9} & 42.9 \\ \bottomrule
\end{tabular}
\caption{\textbf{Comparison of mask sampling strategies.}
We report \rgb-only transfers to ImageNet-1K~\cite{Deng2009ImageNet} classification and ADE20K~\cite{Zhou2017ADE20K} semantic segmentation using {\mmae}s pre-trained with different Dirichlet concentration parameter $\alpha$.
All models were trained for 400 epochs and do \textit{not} use the additional per-patch-standardized \rgb decoder (see Sec.~\ref{sec:pseudo_labeling}).
By $\alpha = \infty$ we denote always sampling an equal number of visible tokens for each tasks.
}
\label{tab:alpha_ablation}
\end{table}

\section{Detailed Taskonomy transfer results} \label{sec:taskonomy_supp}

Table \ref{tab:ablations} compared several baselines by their average rank on eight different Taskonomy~\cite{Zamir2018Taskonomy} downstream tasks.
In this section, we show per-task results of all these baselines.
Table~\ref{tab:taskonomy_multimae} shows detailed results for the ablation on the choice of \mmae pre-training tasks, while Table~\ref{tab:taskonomy_baselines} shows results for the comparison to single-task and multi-task baselines.

Out of these eight Taskonomy tasks, the edges and 2D-keypoints task labels were originally created from RGB images, while the other tasks were rendered from the scanned scene mesh.
A pre-training scheme that includes depth should thus transfer better to the depth-related tasks, such as surface normals.
Indeed, we observe this in Table~\ref{tab:taskonomy_multimae}, where \mmae pre-trained using depth transfer better than MAE or the \rgbs \mmae.
Importantly, additionally including \csemseg{semantic segmentation} along \rgb and \cdepth{depth} in the pre-training does not degrade performance on these tasks.

In Table~\ref{tab:taskonomy_baselines}, we see that \mmae performs similarly to the single-task \crgb{RGB}$\rightarrow$\depth baseline that was trained using full \rgb inputs.
For the single and multi-task baselines, the right choice of pre-training task(s) is crucial, as for example the \crgb{RGB}$\rightarrow$\semseg baselines performs consistently worse than the ones including \cdepth{depth}, as well as the \mmae \rgb-\semseg baseline from Table~\ref{tab:taskonomy_multimae}.

\section{Robustness evaluation on ImageNet} \label{sec:robustness_supp}
We study the robustness of the ImageNet~\cite{Deng2009ImageNet} fine-tuned models by evaluating them on four different ImageNet-like validation sets~\cite{Hendrycks2021ImageNetA, Hendrycks2019ImageNetC, Hendrycks2021ImageNetR, Wang2019ImageNetS} that contain various domain-shifts and corruptions, and we show the results in Table~\ref{tab:imagenet_robustness}.
To that end, we directly use the models that were fine-tuned on ImageNet-1K classification, and evaluate them without any modifications on the respective robustness evaluation datasets.
\mmae performs better than all baselines of the same model size (ViT-B) on ImageNet-R and ImageNet-S.
It also performs better than MAE on ImageNet-C, but falls behind DINO~\cite{Caron2021DINO} and MoCo-v3~\cite{Chen2021MoCoV3}.
On ImageNet-A, \mmae performs worse than DINO and MAE, but better than the supervised and MoCo-v3 baselines.

\begin{table}[htpb]
\centering
\resizebox{1.0\columnwidth}{!}{%
\begin{tabular}{@{}lccccc@{}}
\toprule
Method & IN-1K $\uparrow$ & IN-A $\uparrow$ & IN-C $\downarrow$ & IN-R $\uparrow$ & IN-S $\uparrow$ \\ \midrule
Supervised \cite{Touvron2021DeiT} & 81.8 & 24.2 & 49.7 & 43.5 & 31.4 \\
DINO \cite{Caron2021DINO} & 83.1 & \textbf{35.5} & \textbf{45.5} & 48.1 & 35.4 \\
MoCo-v3 \cite{Chen2021MoCoV3} & 82.8 & 33.2 & \underline{46.2} & 48.4 & \underline{35.6} \\
MAE \cite{He2021MAE} & \textbf{83.3} & \underline{35.1} & 51.6 & \underline{49.3} & 35.5 \\ \midrule
\mmae & \textbf{83.3} & 33.9 & 49.1 & \textbf{50.5} & \textbf{37.1} \\ \bottomrule
\end{tabular}
}
\caption{
\textbf{Robustness evaluation} on ImageNet variants from \textbf{\rgb-only}. 
We report the top-1 accuracy on the IN-1K validation split, as well as robustness evaluations on IN-Adversarial~\cite{Hendrycks2021ImageNetA}, IN-Corruption~\cite{Hendrycks2019ImageNetC} (mean corruption error), IN-Rendition~\cite{Hendrycks2021ImageNetR}, as well as IN-Sketch~\cite{Wang2019ImageNetS}.
}
\label{tab:imagenet_robustness}
\end{table}

\begin{table*}[htpb]
\resizebox{\textwidth}{!}{%
\begin{tabular}{@{}lcccccccc|cc@{}}
\toprule
Method & \multicolumn{1}{l}{\begin{tabular}[c]{@{}l@{}}Curvature\\ ($\cdot 10^2$)\end{tabular}} & \multicolumn{1}{l}{\begin{tabular}[c]{@{}l@{}}Depth\\ ($\cdot 10^2$)\end{tabular}} & \multicolumn{1}{l}{\begin{tabular}[c]{@{}l@{}}Edges\\ ($\cdot 10^3$)\end{tabular}} & \multicolumn{1}{l}{\begin{tabular}[c]{@{}l@{}}Occlusion\\ ($\cdot 10^4$)\end{tabular}} & \multicolumn{1}{l}{\begin{tabular}[c]{@{}l@{}}2D-keypoints\\ ($\cdot 10^4$)\end{tabular}} & \multicolumn{1}{l}{\begin{tabular}[c]{@{}l@{}}3D-keypoints\\ ($\cdot 10^2$)\end{tabular}} & \multicolumn{1}{l}{\begin{tabular}[c]{@{}l@{}}Normals\\ ($\cdot 10^2$)\end{tabular}} & \multicolumn{1}{l|}{\begin{tabular}[c]{@{}l@{}}Reshading\\ ($\cdot 10$)\end{tabular}} & \multicolumn{1}{l}{\begin{tabular}[c]{@{}l@{}}Average \\ loss ($\cdot 10^2$)\end{tabular}} & \multicolumn{1}{l}{\begin{tabular}[c]{@{}l@{}}Average \\ rank\end{tabular}} \\ \midrule
MAE (D2) & 4.455 & 3.651 & 4.608 & 6.237 & 2.736 & 4.585 & 6.189 & 1.120 & 3.828 & 3.75 \\
\rgbd & {\ul 4.249} & {\ul 3.378} & {\ul 4.031} & 6.608 & \textbf{2.440} & {\ul 4.447} & {\ul 6.094} & {\ul 1.051} & {\ul 3.646} & {\ul 2.125} \\
\rgbs & 4.276 & 3.406 & \textbf{3.868} & {\ul 5.939} & 2.615 & 4.467 & 6.139 & 1.067 & 3.678 & 2.625 \\
\rgbds & \textbf{4.236} & \textbf{3.340} & 5.290 & \textbf{5.924} & {\ul 2.590} & \textbf{4.432} & \textbf{6.086} & \textbf{1.040} & \textbf{3.639} & \textbf{1.5} \\ \bottomrule
\end{tabular}
}
\caption{
\textbf{Taskonomy transfer results} using \mmae models pre-trained on a \textbf{varying number of modalities}, where the pre-training modalities are the same as the target tasks.
Downstream transfers are trained from \rgb-only.
All models were pre-trained for 400 epochs.
We report L1 losses ($\downarrow$) and indicate with \textbf{bold} and {\ul underline} the best and second-best results, respectively.
}
\label{tab:taskonomy_multimae}
\end{table*}

\begin{table*}[htpb]
\resizebox{\textwidth}{!}{%
\begin{tabular}{@{}lllllllll|ll@{}}
\toprule
Method & \begin{tabular}[c]{@{}l@{}}Curvature\\ ($\cdot 10^2$)\end{tabular} & \begin{tabular}[c]{@{}l@{}}Depth\\ ($\cdot 10^2$)\end{tabular} & \begin{tabular}[c]{@{}l@{}}Edges\\ ($\cdot 10^3$)\end{tabular} & \begin{tabular}[c]{@{}l@{}}Occlusion\\ ($\cdot 10^4$)\end{tabular} & \begin{tabular}[c]{@{}l@{}}2D-keypoints\\ ($\cdot 10^4$)\end{tabular} & \begin{tabular}[c]{@{}l@{}}3D-keypoints\\ ($\cdot 10^2$)\end{tabular} & \begin{tabular}[c]{@{}l@{}}Normals\\ ($\cdot 10^2$)\end{tabular} & \begin{tabular}[c]{@{}l@{}}Reshading\\ ($\cdot 10$)\end{tabular} & \begin{tabular}[c]{@{}l@{}}Average \\ loss ($\cdot 10^2$)\end{tabular} & \begin{tabular}[c]{@{}l@{}}Average \\ rank\end{tabular} \\ \midrule
\crgb{RGB}$\rightarrow$\depth & {\ul 4.251} & \textbf{3.222} & 7.038 & \textbf{5.914} & {\ul 2.790} & {\ul 4.458} & \textbf{5.960} & \textbf{1.013} & \textbf{3.602} & {\ul 1.625} \\
\crgb{RGB}$\rightarrow$\semseg & 4.314 & 3.666 & 7.206 & 6.051 & 3.029 & 4.595 & 6.843 & 1.155 & 3.973 & 4 \\
\crgb{RGB}$\rightarrow$\depthsemseg & 4.266 & 3.465 & {\ul 6.745} & 5.949 & 2.899 & 4.510 & 6.264 & 1.080 & 3.759 & 2.875 \\
\mmae & \textbf{4.236} & {\ul 3.340} & \textbf{5.290} & {\ul 5.924} & \textbf{2.590} & \textbf{4.432} & {\ul 6.086} & {\ul 1.040} & {\ul 3.639} & \textbf{1.5} \\ \bottomrule
\end{tabular}
}
\caption{
\textbf{Taskonomy transfer results} comparing pre-trained single-task and multi-task baselines (pre-trained using \textbf{\textit{non-masked}} \rgb-only inputs) against the \rgbds \mmae.
Downstream transfers are trained from \rgb-only.
All models were pre-trained for 400 epochs.
We report L1 losses ($\downarrow$) and indicate with \textbf{bold} and {\ul underline} the best and second-best results, respectively.
}
\label{tab:taskonomy_baselines}
\end{table*}

\section{Comparison of MAE variants} \label{sec:mae_diff_supp}

In Section \ref{sec:rgb_transfers}, we compare \mmae to a pre-trained MAE with a decoder of depth 8, following the best-performing setting described in \cite{He2021MAE}. However, as our \mmae uses shallower and narrower decoders, we also pre-train MAE with a decoder of similar depth (2) and width (256). We compare these two MAE versions in Table \ref{tab:mae-comparison}. We find that while these two models perform comparably on ImageNet-1K classification, as reported in \cite{He2021MAE}, using a deeper decoder leads to a stark increase in performance for all other tasks. Given the benefits of a larger decoder for MAE, it stands to reason that \mmae could also benefit from using wider and deeper decoders, even though that would significantly increase pre-training time. 

Furthermore, it has been observed that MAE models pre-trained using the official PyTorch~\cite{Paszke2019PyTorch} implementation (such as ours) do not exactly match the results of a MAE trained using the original (and unavailable) TensorFlow~\cite{tensorflow2015-whitepaper} implementation\footnote{A discussion about the reproducibility issues  of MAE can be found at: \href{https://github.com/facebookresearch/mae/issues/30}{https://github.com/facebookresearch/mae/issues/30}}. Therefore, we also report results using model weights from the TensorFlow implementation to assess the impact of the codebase on transfer performance. We observe minor differences in transfer performance, with the original TensorFlow implementation slightly outperforming the PyTorch implementation on all tasks.

\begin{table}[htpb]
\centering
\resizebox{1.0\columnwidth}{!}{%
\begin{tabular}{@{}lccccc@{}}
\toprule
Method & \small IN-1K (C) & \small ADE20K (S) & \small Hypersim (S) & \small NYUv2 (S) & \small NYUv2 (D) \\ \midrule
MAE (D2, PyTorch) & \underline{83.3} & 43.3 & 34.1 & 46.9 & 83.7 \\
MAE (D8, PyTorch)  & \underline{83.3} & \underline{46.2} & \underline{36.5} & \underline{50.1} & \underline{85.1} \\
MAE (D8, TensorFlow) & \textbf{83.6} & \textbf{46.5} & \textbf{37.1} & \textbf{50.9} & \textbf{85.4} \\
\bottomrule
\end{tabular}
}
\caption{\textbf{Comparison of MAE variants.}
We report the top-1 accuracy ($\uparrow$) on ImageNet-1K~\cite{Deng2009ImageNet} (IN-1K) classification (C), mIoU ($\uparrow$) on ADE20K~\cite{Zhou2017ADE20K}, Hypersim~\cite{Roberts2021Hypersim}, and NYUv2~\cite{Silberman2012NYUv2} semantic segmentation (S), as well as $\delta_1$ accuracy ($\uparrow$) on NYUv2 depth (D). Text in \textbf{bold} and \underline{underline} indicates the first and second-best results, respectively. All models are pre-trained for 1600 epochs. D2 = Decoder of depth 2 and width 256. D8 = Decoder of depth 8 and width 512.
}
\label{tab:mae-comparison}
\end{table}

\section{Comparison of pre-training time } \label{sec:runtime_supp}

We report the pre-training epoch time in Table \ref{tab:runtime}. By using shallow decoders, the training time of \mmae is comparable to MAE (with a decoder of depth 8) despite having twice the amount of unmasked tokens and multiple decoders. Note that removing masked tokens from the encoder, as proposed by MAE, is crucial in enabling pre-training on multiple dense modalities. 

\addtolength{\tabcolsep}{3pt}    
\begin{table}[htpb]
\centering
\resizebox{1.0\columnwidth}{!}{%
\begin{tabular}{@{}llcc@{}}
\toprule
Method &  \small Encoder & \small Num. unmasked &  \small Epoch time (mins) \\ \midrule
MAE (D2) & ViT-B & 49 & 2.7  \\
MAE (D8) & ViT-B & 49 &  5.0 \\
\midrule
\mmae & ViT-B & 98 & 6.0  \\
\mmae, w/ \texttt{[M]} & ViT-B& 98 & 43.3 \\
\bottomrule
\end{tabular}
}
\caption{\textbf{Pre-training time comparison.} Pre-training epoch time for MAE~\cite{He2021MAE} and \mmae on ImageNet-1K~\cite{Deng2009ImageNet}. We train with 8 Nvidia A100 GPUs and use PyTorch with automatic mixed precision enabled. D2 = Decoder of depth 2 and width 256. D8 = Decoder of depth 8 and width 512. w/ \texttt{[M]} = Mask tokens also given to the ViT-B encoder.
}
\label{tab:runtime}
\end{table}
\addtolength{\tabcolsep}{-3pt}

\section{Additional visualizations} \label{sec:viz_supp}
\begin{figure*}[t]
\centering
\includegraphics[width=0.95\textwidth]{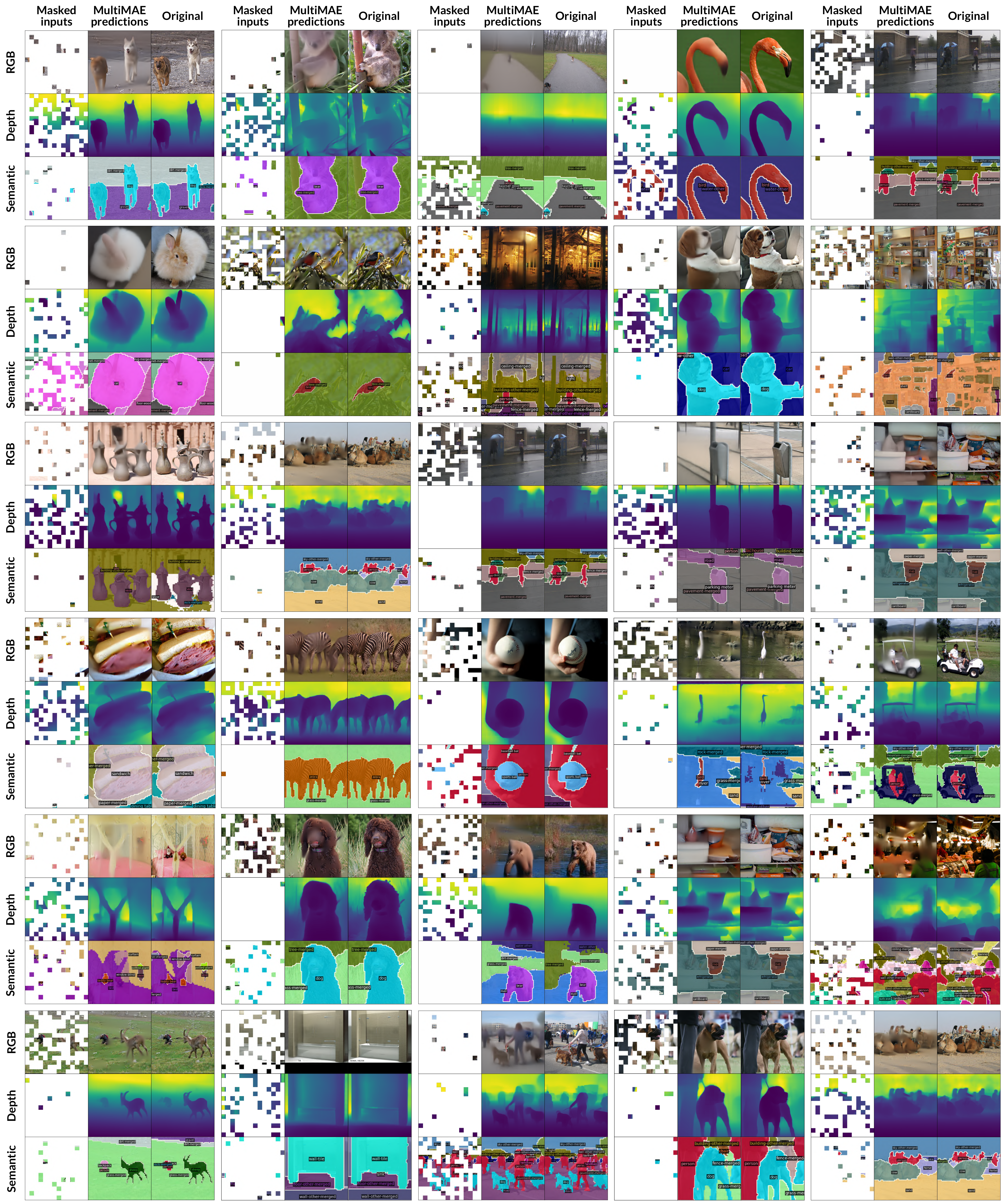}
\caption{
   \textbf{ \mmae predictions on ImageNet-1K validation set samples.} 
    98 visible patches were sampled using Dirichlet concentration parameter $\alpha = 1$.
}
\label{fig:random_samples_supp}
\centering
\vspace{-1em}
\end{figure*}

\begin{figure*}[t]
\centering
\includegraphics[width=0.95\textwidth]{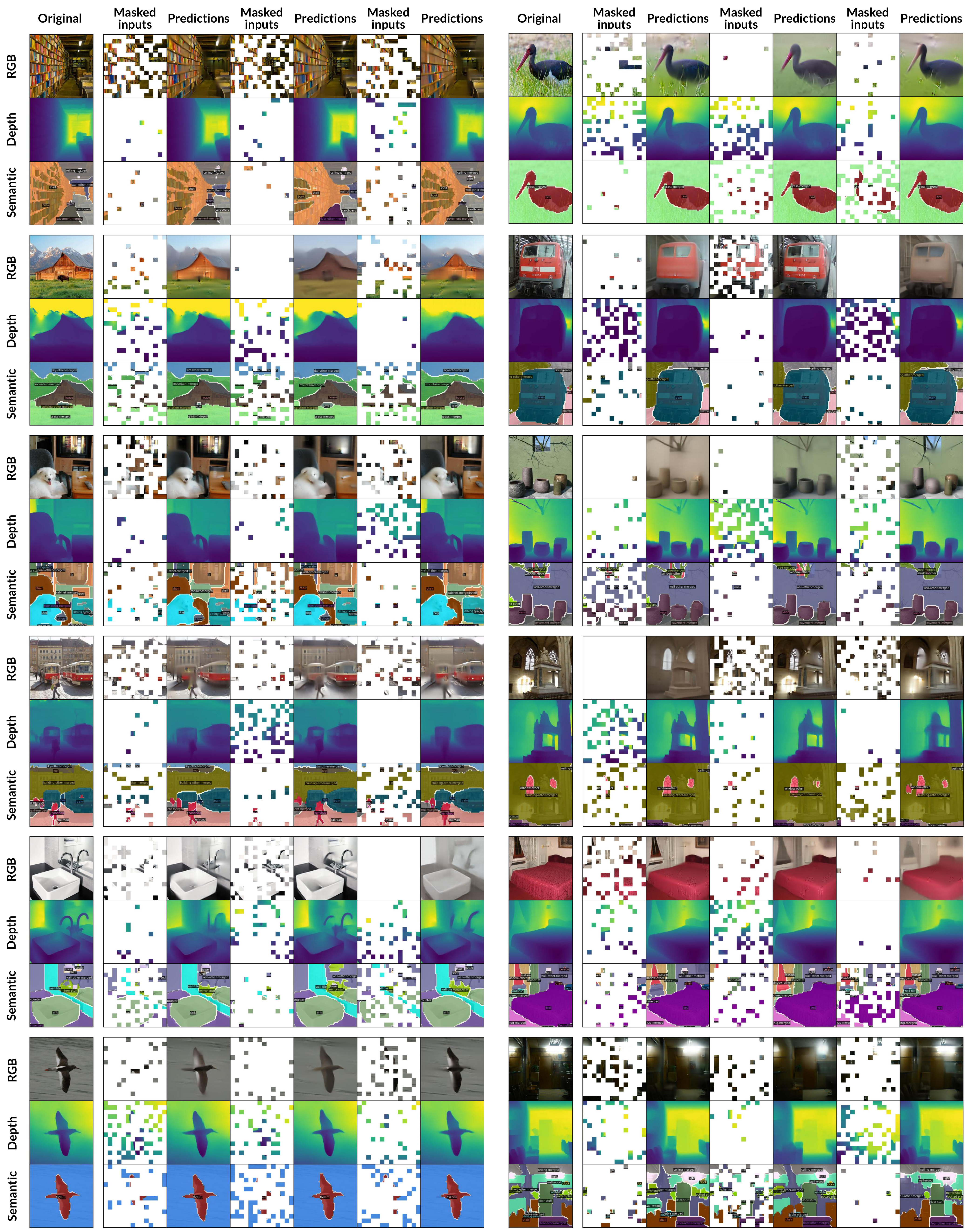}
\caption{
    \textbf{\mmae predictions on ImageNet-1K validation set samples.} 
    98 visible patches were sampled using Dirichlet concentration parameter $\alpha = 1$.
    For each image, we sample three random masks.
}
\label{fig:random_variants}
\centering
\vspace{-1em}
\end{figure*}

Figure~\ref{fig:random_samples_supp} shows more visualizations on ImageNet-1K~\cite{Deng2009ImageNet} validation set images.
For all examples, 98 visible patches were sampled using Dirichlet concentration parameter $\alpha = 1$.
Figure~\ref{fig:random_variants} further shows predictions where we sample three random masks for each image.

\end{document}